\pdfoutput=1

\documentclass[11pt]{article}

\usepackage[]{acl}

\usepackage{times}
\usepackage{latexsym}

\usepackage[T1]{fontenc}

\usepackage[utf8]{inputenc}

\usepackage{microtype}

\usepackage{url}
\usepackage{times}
\usepackage{latexsym}

\usepackage{multicol}
\usepackage{multirow}
\usepackage{graphicx}
\usepackage{verbatim}
\usepackage{makecell}
\usepackage{braket}
\usepackage{hhline}
\usepackage{xfrac}
\usepackage{amsmath,bm,amssymb}
\usepackage[linesnumbered,ruled]{algorithm2e}
\usepackage{xcolor}
\definecolor{Green}{rgb}{0.0, 0.5, 0.0}
\definecolor{Amethyst}{rgb}{0.6, 0.4, 0.8}
\usepackage{colortbl}
\title{A Meta-Evaluation of Faithfulness Metrics \\ for Long-Form Hospital-Course Summarization }

\author{
Griffin Adams \\
  Columbia University \\
  \texttt{griffin.adams@columbia.edu} \\\And
  Jason Zucker \\
  Columbia University Irving Medical Center \\
  \texttt{jz2700@cumc.columbia.edu}
\AND
  No\'emie Elhadad \\
  Columbia University \\
  \texttt{noemie.elhadad@columbia.edu} \\ }

\begin{document}
\maketitle
\begin{abstract}

Long-form clinical summarization of hospital admissions has real-world significance because of its potential to help both clinicians and patients. The faithfulness of summaries is critical to their safe usage in clinical settings. To better understand the limitations of abstractive systems, as well as the suitability of existing evaluation metrics, we benchmark faithfulness metrics against fine-grained human annotations for model-generated summaries of a patient's Brief Hospital Course. We create a corpus of patient hospital admissions and summaries for a cohort of HIV patients, each with complex medical histories. Annotators are presented with summaries and source notes, and asked to categorize manually highlighted summary elements (clinical entities like conditions and medications as well as actions like "following up") into one of three categories: ``Incorrect,'' ``Missing,'' and ``Not in Notes.'' We meta-evaluate a broad set of proposed faithfulness metrics and, across metrics, explore the importance of domain adaptation (e.g. the impact of in-domain pre-training and metric fine-tuning), the use of source-summary alignments, and the effects of distilling a single metric from an ensemble of pre-existing metrics. Off-the-shelf metrics with no exposure to clinical text correlate well yet overly rely on summary extractiveness. As a practical guide to long-form clinical narrative summarization, we find that most metrics correlate best to human judgments when provided with one summary sentence at a time and a minimal set of relevant source context.
\end{abstract}


\section{Introduction} \label{sec:introduction}

A significant factor for clinician burnout is the Electronic Health Record (EHR), the information overload it produces, and the documentation burden it requires \citep{shanafelt2016relationship, moy2021measurement}. A study of US physicians revealed that doctors spent 27\% of working hours with patients and nearly 50\% of their time on EHR and desk work, in addition to 1-2 hours at night, spent mostly on documentation \citep{sinsky2016allocation}. Clinician burnout can have damaging consequences not only for clinicians \citep{national2019taking}, due to, among other factors, increased rates of depression \citep{maslach2016understanding} and interrupted work-life balance \citep{kroth2019association}), but also patients, due to increased risk of clinical errors \citep{salvagioni2017physical, panagioti2018association}.


In the inpatient setting, the Discharge Summary \citep{kind2008documentation} is a particularly tedious and time-consuming note to write \citep{chan2014improving}. Yet, it is a critical piece of documentation. Written at the end of a patient's hospital admission, the Discharge Summary ensures continuity of care \citep{kripalani2007deficits, o2009creating}. Its timely availability has been shown to have a direct impact on patient quality of care, including the rate of hospital readmission \citep{van2002effect}. A key mandatory section of the Discharge Summary is the ``Brief Hospital Course,'' which, in a paragraph of variable-length, recounts in a narrative form the events occurred during the patient stay, and why they happened. Composing the hospital-course summary is a cognitively difficult task for clinicians. They must review a high number of clinical notes and reports entered during the patient stay and synthesize them into a long paragraph. It is even more challenging when an admission is complex, which is often the case for patients with comorbidities or chronic conditions.

Automated summarization techniques can support clinicians in this difficult task. An automatically generated hospital course summary can act as a first draft for a clinician and ensure that the critical elements of the patient stay are not missed in the potentially overwhelmingly large amount of notes produced during the patient stay. Generating a high-quality hospital course narrative is difficult and ensuring its faithfulness is paramount: this requires synthesizing and fusing information from diverse note types, while remaining consistent: adhering to temporal constraints, providing sufficient context to avoid misleading patient characterizations, and even resolving source note errors. 

Long-form abstractive summarization is an active topic of research in the general domain \citep{guo2021longt5,phang2022investigating}, yet most faithfulness metrics have been developed on shorter datasets~\cite{factcc,durmus-etal-2020-feqa,wang-etal-2020-asking,deng-etal-2021-compression,yuan2021bartscore,laban-etal-2022-summac,ribeiro-etal-2022-factgraph}. In the clinical domain, there are additional open questions, including the performance of modern summarization models and whether existing evaluation metrics are truly reflective of clinical quality. In this paper, we examine the performance of an established long-form abstractive summarization model on the task of hospital course summarization, as well as the quality of existing faithfulness metrics when compared to clinicians' judgments.

To this end, we fine-tune a long-range transformer (Longformer Encoder-Decoder (LED) \citep{beltagy2020longformer} on a large dataset of Hospital Course summaries, pertaining to all in-patient hospital admissions at a large healthcare institution (Columbia University Irving Medical Center in New York City) from 2010-2014 \citep{adams-etal-2021-whats}. On a held-out set of admissions for patients from the HIV clinic \citep{levy2020towards}, we rely on expert (clinicians) to collect fine-grained faithfulness annotations of LED summaries based on the clinical notes written before discharge.

We then meta-evaluate a large set of existing summarization evaluation metrics (including BARTScore \citep{yuan2021bartscore}, BERTScore \citep{zhang2019bertscore}, Entailment-based CTC \citep{ctc} and SummaC \citep{laban-etal-2022-summac}) by measuring their correlation to human annotations. Since these metrics were mostly developed on single document general-domain corpora, we identify three key dimensions pertinent to adaptation to long-form clinical summarization: domain adapation (pre-training and metric fine-tuning), length of inputs, and length of outputs. For length-based dimensions, we explore the impact of source-summary alignments and summary granularity (sentence-level versus summary-level). We find that metrics tend to correlate best with human annotations when provided summary sentences one at a time, and when only the most relevant content (high precision source-summary alignments) is provided. We see limited benefits from domain adaptation with respect to simple correlation analysis, yet we attribute much of this to the abstractiveness of the references on which metrics are tuned. When filtering for abstractive subsets of the annotation set, domain adaptation starts to outperform off-the-shelf variants. In-domain adaptation of metrics will likely be critical given the observed abstractiveness of summaries from LLMs \citep{goyal2022news}. 

Rather than adapt metrics to clinical text by training on references, we find it advantageous to learn directly from system summaries. We use an ensemble of our baseline metrics to produce a pseudo faithfulness score on system summaries and distill a metric from these noisy ground-truth labels. Our distilled metric has a higher correlation than baseline metrics to expert annotation labels.

Our contributions are: \textbf{(1)} We collect fine-grained faithfulness annotations for the the task of hospital-course summarization, which contains substantially longer inputs than previous clinical annotation efforts; \textbf{(2)} We benchmark existing faithfulness metrics against these annotations, as well as explore practical considerations of adapting general domain metrics to long-form clinical narratives; \textbf{(3)} We analyze the confounding role of extractiveness and show how a simple statistic (unigram coverage) can be complementary to other metrics, including a metric distilled from an ensemble of other metrics.


\section{Related Work} \label{sec:related-work}

\paragraph{Faithfulness Metrics.} Metrics to assess faithfulness can be roughly distilled into the following categories: QA-based \citep{wang-etal-2020-asking, fabbri-etal-2022-qafacteval, durmus-etal-2020-feqa}, entailment based metrics from NLI \citep{falke-etal-2019-ranking} or synthetic data \citep{kryscinski-etal-2020-evaluating, deng-etal-2021-compression, utama-etal-2022-falsesum}, fact-based, reference-free overlap \citep{goodrich2019assessing}, and those which directly learn from human judgments \citep{ribeiro-etal-2022-factgraph} (similar to BLEURT \citep{sellam-etal-2020-bleurt} approach for machine translation). Most of these metrics have been developed on single document news summarization datasets, such as CNN / DailyMail \citep{HermannKGEKSB15, see-etal-2017-get} and Xsum \citep{narayan-etal-2018-dont}. Faithfulness metrics proposed for clinical summary evaluation have typically come from the overlap category and focus on concept alignment between summaries and the source input \citep{zhang-etal-2020-optimizing, tang-etal-2022-echogen}.

\paragraph{Human Faithfulness Evaluation.} Assessing faithfulness is a challenging task to automate with metrics \citep{bhandari-etal-2020-evaluating}, which underscores the importance of collecting high-quality human evaluation annotations \citep{lux-etal-2020-truth, wang-etal-2020-asking, kryscinski-etal-2020-evaluating, maynez-etal-2020-faithfulness, huang-etal-2020-achieved, fabbri-etal-2021-summeval, pagnoni-etal-2021-understanding, goyal-durrett-2021-annotating, cao-wang-2021-cliff, cao-etal-2022-hallucinated}. Additionally, given the relatively small size of each separate evaluation, it can be useful for training and/or meta-evaluation to aggregate them into larger benchmark datasets \citep{fabbri-etal-2021-summeval, laban-etal-2022-summac}.

Based on low inter-annotator agreements for summary-level faithfulness annotations \citep{lebanoff-etal-2019-analyzing, factcc}, recent work has focused more on fine-grained annotations at the entity \citep{cliff, cao-etal-2022-hallucinated}, sentence \citep{pagnoni-etal-2021-understanding}, and span level \citep{maynez-etal-2020-faithfulness}. These studies tend to have higher annotator agreement and allow for a better understanding of the typology of error distributions across datasets and systems. Sophisticated error taxonomies are generally formulated by examining system outputs (e.g., card-sorting exercises \citep{lux-etal-2020-truth}) and tend to demarcate error types on two fronts: where the error is located (broken down by syntactic roles) and where it \emph{likely} comes from (intrinsic or extrinsic). \citet{zhang2022extractive} challenges the notion that extractive summaries are consistent by analyzing inter-sentence discourse.

\paragraph{Evaluation of Clinical Note Summarization.}

\citet{moen2014evaluation} evaluate extractively generated Discharge Summaries based on content criteria guidelines and benchmark ROUGE against these coverage-focused annotations. Much of the recent work on human evaluation of clinical summarization has focused on self-contained, single-document tasks: including radiology report summarization \citep{macavaney2019ontology, zhang-etal-2020-optimizing} and echocardiogram conclusions \citep{tang-etal-2022-echogen}. For these shorter tasks, summary-level assessments are collected, in the form of pairwise ranking \citep{tang-etal-2022-echogen} or point-wise assessments \citep{macavaney2019ontology} on a Likert Scale. \citet{moramarco-etal-2021-towards} examine brief descriptions of SOAP notes for mock patient encounters (MTSamples\footnote{\url{https://mtsamples.com}}, and compare fact-based overlap between reference and system-generated summaries.

Most closely related to our work, \citet{moramarco-etal-2022-human} perform a human evaluation on a more self-contained, conditional clinical note generation task: generating a SOAP note from consultation transcripts. They rely on a dataset of mock patient-doctor conversations and corresponding SOAP notes from \citet{korfiatis2022primock57}. Annotators were asked to post-edit notes to correct errors, as well as manually highlight spans with incorrect or omitted information.  Automatic metrics were then benchmarked against post-editing time, as well as the number of incorrect and omitted spans. Our work differs as we define a typology of errors with more categories, consider more diverse faithfulness metrics, and, because our data includes much longer clinical narratives, explore the impact of using source-summary alignments and different summary granularities (sentence-level versus full).

\begin{table}[h]
\centering
\small
\begin{tabular}{l|c|cc|cc}
\multirow{2}{*}{\texttt{Split}} & \multirow{2}{*}{\texttt{\#}} & \multicolumn{2}{c}{\texttt{Source}} & \multicolumn{2}{c}{\texttt{Reference}}  \\
& & Notes & Tokens & Sents & Tokens \\ \hline
Train - Full & 82k & 41 & 18.4k & 11.6 & 207 \\  
Train - HIV & 2.7k & 40 & 19.1k & 12.5 & 243 \\ 
Eval - HIV & 29 & 24 & 11.7k & 12.1 & 211 \\ 
\end{tabular}
\caption{Data Statistics for training the summarization LED model (Full Train), the subset used for in-domain \emph{evaluation metric} training, as well as the subset of the test set used for human evaluation (Annot.). } \label{tab:data-stats}
\end{table}

\section{Data}

The data is comprised of clinical notes from the Electronic Health Record (EHR) for in-patient admissions between 2010 and 2014 at a large metropolitan hospital \citep{adams-etal-2021-whats}.

\paragraph{Training Data.} We show training data statistics in the first row of Table \ref{tab:data-stats}. We delineate between the full training set, which is used to train the summarization models and the subset of the training set which is used for fine-tuning evaluation metrics in-domain. The subset filters for HIV patients which mirrors the filtering done to produce the human evaluation cohort (discussed directly below).

\paragraph{Human Evaluation Cohort.} The training set comprises both HIV and non-HIV patients while the human annotation test set is solely HIV. We choose HIV patients as they typically have multiple co-morbidities and, concomitantly, complex hospital courses \citep{gallant2017comorbidities}. We first filter the test set for patients admitted to the HIV clinic (10k to 339 admissions) \citep{levy2020towards}. From this HIV-specific cohort, we remove outliers: the top and bottom ten percent by number of source notes, and do the same for the summary reference length. The admissions with the fewest notes tend to cover cases in which notes are missing and, as such, are difficult to annotate. Removing the longest examples (source notes and reference length) filters out a long tail of examples for which obtaining human annotations would be too time consuming. After filtering, we end up with 212 admissions. From this set, we bin the summaries by extractiveness (density) into deciles, similarly to \citet{bhandari-etal-2020-evaluating}, and sample an equal number from each decile to ensure diversity in summaries for annotation. We sample from each bin and end up with 29 summaries for annotation (245 sentences), based on a total of 703 source notes.

\paragraph{Generating Summaries for Annotation.} At a high-level, we fine-tune a Transformer Encoder-Decoder with sparse attention (Longformer Encoder-Decoder (LED) \citep{beltagy2020longformer}). The LED handles inputs up to 16,384 tokens. To fit all inputs (the average input length from Table \ref{tab:data-stats} is $18.4k$), we train a simple bi-LSTM model to rank each sections and, during inference, retain the top 100 sections. Filtering and fine-tuning details and hyper-parameters are provided in Appendix \ref{app:led}.

\begin{figure*}[t]
\centering
\includegraphics[width=\linewidth]{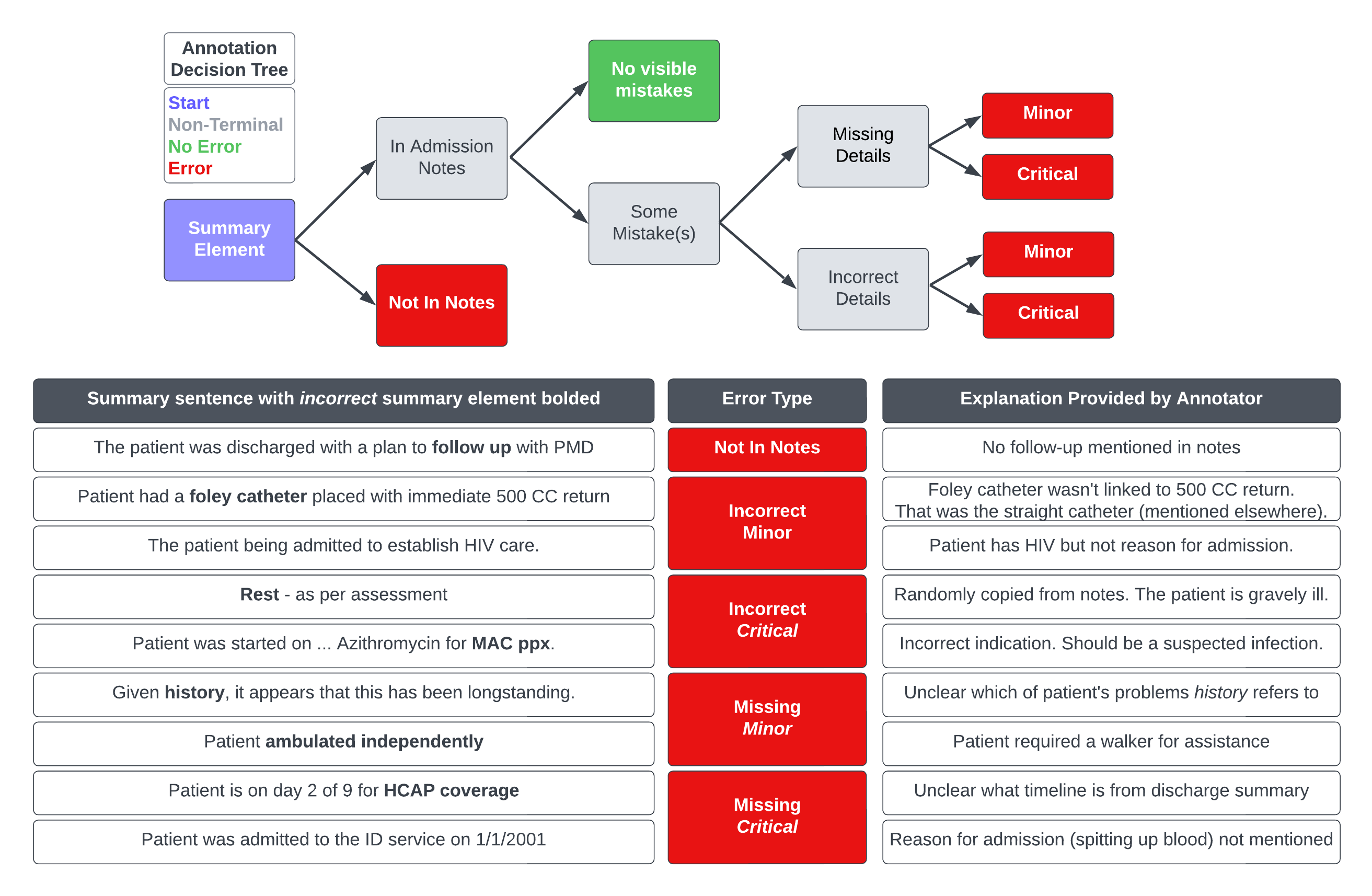}
\caption{Annotation Decision Tree with real, but modified, examples for each error type. Examples have been modified to removing any protected health information (PHI) and change all numbers (e.g., dates). } 
\label{fig:annotation-tree}
\end{figure*}

\section{Collecting Annotations} \label{sec:annotation-protocol}

At a high-level, the annotation task consisted of assigning an error category (or No Error) to each Summary Element (defined below) in a system output, based solely on clinical knowledge and all patient's clinical notes from the hospital admission.

\paragraph{Summary Elements.} As in other faithfulness work \citep{goyal-durrett-2021-annotating}, we decided to collect fine-grained annotations and experimented with different granularities while piloting the study. We found that entities (used in \citet{cao-etal-2022-hallucinated}) were too granular, noisy, and incomplete on clinical notes. Syntactic parses were unreliable on our text as well. On the other hand, sentence-level annotation \citep{wang-etal-2020-asking, durmus-etal-2020-feqa, pagnoni-etal-2021-understanding} was insufficiently fine-grained given the length and information density of many sentences. As such, the authors of the paper manually extracted \texttt{Summary Elements} (SE), which consist of standard medical concepts and actions, as well as compound concepts. Standard medical concepts included Disorders, Medications, Procedures, and Treatments, while actions encapsulate phrases such as ``discharged to home'' and ``plans to follow up''. When sensible, we merged compound entities into a single \textbf{SE}: ``alkanization of urine''.

\paragraph{Error Categories.} For each SE, annotators were asked to identify and categorize errors. As represented as a decision tree in Figure \ref{fig:annotation-tree}, annotators were first asked to confirm whether or not the summary element is ``hallucinated'': \texttt{Not in Notes}. If the SE can be found in the notes, they either deem it correct: \texttt{No visible mistakes} or denote an inconsistency in its usage. For these intrinsic-focused errors, we delineate between \texttt{Incorrect Details} and \texttt{Missing Details}. A SE has \texttt{Incorrect Details} if it can be found in the source notes yet contains information that does not reflect what is written in the notes. This category encapsulates numerical errors (dosages, dates), mis-representations of symptoms (``afebrile'' is incorrect if patient had a fever), fusion errors (an incorrect indication for a drug), among others. An SE has a \texttt{Missing Details} error if the summary omits important information about the SE, which could lead to misleading conclusions being formed about the patient’s true hospital course. \texttt{Missing Details} is grounded on a specific SE and thus less open-ended than previously defined ``omission'' errors \citep{huang-etal-2020-achieved, moramarco-etal-2022-human}.

\paragraph{Severity of Errors.} For \texttt{Incorrect} and \texttt{Missing}, as in \citet{moramarco-etal-2022-human}, we ask annotators to distinguish between Minor and Critical errors. We provide annotators with examples of both kinds of errors and define \texttt{Critical} as a mistake which could negatively impact the patient's present and future treatment. Minor is an exclusionary category defined as ``Not Critical''.

\paragraph{Annotators.} We recruited $6$ clinical practitioners, with IRB-approved access to the patient data, to annotate the summaries in \texttt{Eval - HIV}. Each annotator was compensated at a rate of \$30 / hour. $4/6$ of the annotators self-identify as female, with the other two as male. $4/6$ self-identify as ``White'', and $1$ each as ``Black or African'' and ``Other''. $2$ annotators are attending physicians, $3$ are in medical residency, and $1$ is a fellow. They have a combined 25 years of medical practice. Each expert annotated summaries for a minimum of one hour at the same computer and location with the guidance of the authors of the paper, who were available for technical guidance and clarification questions regarding the annotation protocol. Collectively, the task was carried out over $\sim 10$ hours across 4 days.

\paragraph{Description of Interface.} We develop a custom annotation interface within Prodigy \citep{prodigy}. The interface presented each annotator with one summary at a time. For viewing ease, summaries were split such that one sentence was shown per line. Summary Elements (SE) were highlighted and annotation of non-SE spans prohibited.  For each SE, annotators would select the appropriate erorr category (or No Error) and then either double click or highlight the SE span. On a separate browser page, we displayed the source notes for the patient visit, which were hosted locally on a custom, light-weight app. The left-hand side of the full-text notes display showed section headers and free text for each note. Notes were sorted by date and annotators could search for a note by its title on a drop-down menu. Section headers were indexed and searchable to allow for efficient navigation of long notes. On the right hand side of the webpage, we enabled free-text search across notes. Each note was pre-indexed such that all mentions of matching search terms across notes could be quickly surfaced. We extracted all concepts with CLAMP NLP, highlighted them in the interface, and allowed for annotators to trigger a concept-based search query by double-clicking on the concept span in the note.

\section{Error Analysis} \label{sec:error-analysis}

\begin{table}[h]
\centering
\small
\setlength{\tabcolsep}{2pt}
\begin{tabular}{l|cc|c}
& \texttt{\textbf{\makecell{Per \\ Summary}}} & \texttt{\textbf{\makecell{Per \\ Sent}}} & \textbf{\texttt{\makecell{\% of \\ All SE}}} \\\hline
All Summary Elements (SE) & 27.10 &3.21 & - \\ \hline
Incorrect SE & 2.86 & 0.34 &11\% \\
Missing SE & 0.93 & 0.11 &3\% \\
Not In Notes SE & 1.03 & 0.12 & 4\% \\ \hline \hline
\emph{Any} Mistake SE & 4.83 &0.57 & 18\% \\
\end{tabular}
\caption{Statistics on Clinician-Annotated Summary Elements (SE), broken down across error categories. }
\label{tab:error-distribution}
\end{table}

\begin{figure}[h]
\centering
\includegraphics[width=\linewidth]{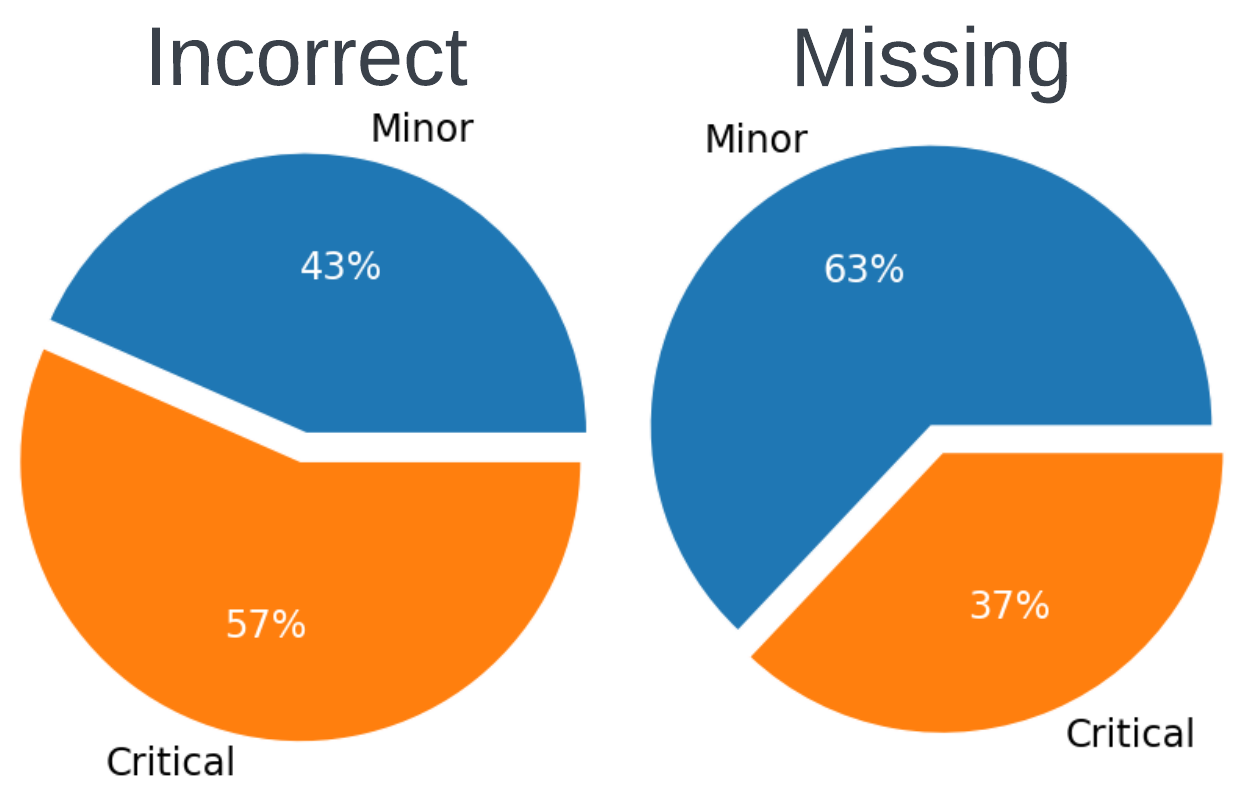}
\caption{Breakdown of errors deemed by clinicians as Minor versus Critical (potentially impacting patient care) for two error types: \texttt{Incorrect} and \texttt{Missing}. } 
\label{fig:critical-pie}
\end{figure}

\begin{figure}[h]
\centering
\includegraphics[width=\linewidth]{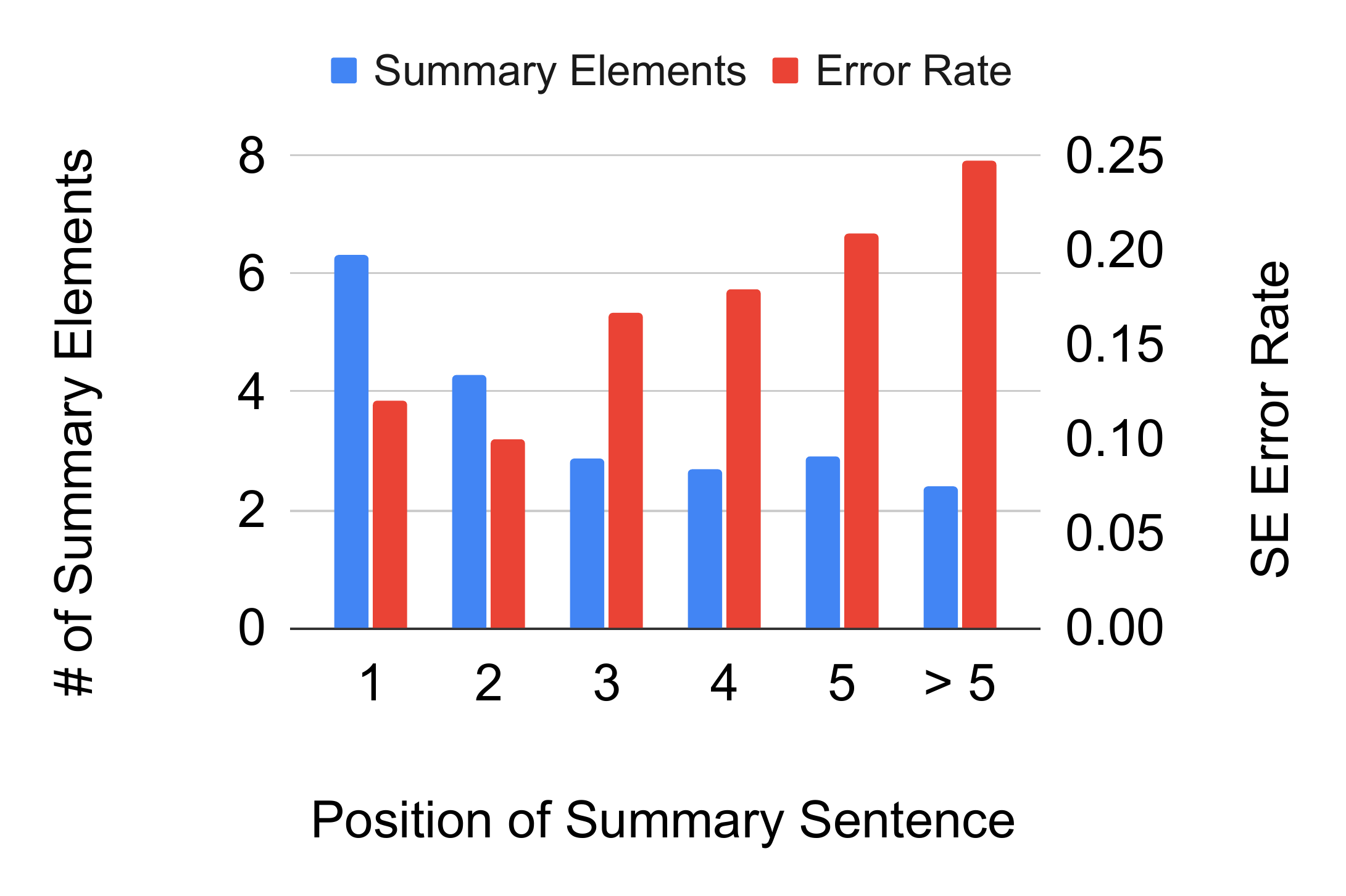}
\caption{Increasing error rate as summary length increases. There are more SEs at the beginning of summaries, which tend to involve longer sentences and many cover lists of diagnoses for the patients (HPI). } 
\label{fig:length}
\end{figure}

\paragraph{Distribution of Errors.} Table \ref{tab:error-distribution} shows the number of SE per summary and per sentence, as well as the breakdown of SE into each error category. $18\%$ of SEs are marked as having \emph{Any mistake}, of which the predominant category is \texttt{Incorrect} ($11\%$ versus $3\%$ and $4\%$ for \texttt{Missing} and \texttt{Not in Notes}). In Table \ref{tab:error-distribution}, Minor and Critical are lumped together and contribute equally to the counts.

\paragraph{Qualitative Analysis.} As shown in Figure \ref{fig:annotation-tree}, incorrect errors often result from improper fusion of concepts: (``foley catheter'' with ``500 CC return'', ``Azithromycin'' with ``MAC ppx'', and ``admitted'' with ``HIV care''). Incorrect errors can also be perfectly extractive. ``Rest - as per assessment'' is copied verbatim from a previous note, yet is incorrect because, at the time of discharge, the patient is actually gravely ill, which contradicts the recommendation. \texttt{Missing Errors} are also quite extractive (see analysis in \S \ref{sec:meta-by-type}) and tend to occur from the reverse problem: insufficient fusion. The model fails to fuse related concepts when they appear in different contexts. Specifically, the model fails to make the following links: use of a ``walker'' is relevant to his ``ambulat[ion]'', that the ``HCAP coverage'' duration should be related to the note timestamp, and that ``admitted to ID service'' should be linked to the reason for admission---``spitting up blood''.

\paragraph{Severity of Errors.} Figure \ref{fig:critical-pie} breaks down error severity for \texttt{Incorrect} and \texttt{Missing}. The majority of \texttt{Incorrect} errors were marked as \textbf{Critical} ($57\%$), whereas a minority for \texttt{Missing} ($37\%$). As implicated by Figure \ref{fig:annotation-tree}, the difference between Critical and Minor errors is very subtle. Typically, the justifications for each, as provided by the annotators, were highly specific to the patient in question. This is interesting as it represents a non-standard definition of faithfulness, one which is more tightly connected to salience, and is grounded on a more holistic view of the patient journey.


\paragraph{Impact of Position in Summary.} Similarly to degeneration in unconditional generation tasks \citep{holtzman2019curious}. we can measure whether or not quality (as measured by faithfulness) declines at different summary positions. Figure \ref{fig:length} plots the percentage of SE marked with any error by the sentence position in the summary. A clear trend emerges of an increasing error rate as summaries grow longer. This may point to a task-agnostic factor: scaling limitations from full self-attention within the decoder, or task-specific factors: a shift in topics. Figure \ref{fig:length} shows the overall number of SEs decreasing by sentence position. From qualitative analysis, we, in fact, observe a topic shift: from dense history of present illness history recounting (diagnosis-heavy) to concise descriptions of procedures and, finally, any post-discharge instructions. 

\section{Evaluation Metrics}


\subsection{Task-Specific Concerns.} \label{sec:challenges}

Broadly speaking, we identify three high-level challenges for evaluating long-form clinical summaries, which are distinct from those faced when evaluating single-document new summaries: \textbf{(1) Domain Adaptation}, \textbf{(1) Long Outputs}, \textbf{(3) Long Inputs}. 

\paragraph{Domain Adaptation.} The first challenge relates to adapting metrics, typically trained and used on general domain data, to clinical text. We cannot adapt all metrics, especially metrics \citep{sellam-etal-2020-bleurt, ribeiro-etal-2022-factgraph} which directly learn from news summary annotation benchmarks \citep{wang-etal-2020-asking, pagnoni-etal-2021-understanding, fabbri-etal-2021-improving, laban-etal-2022-summac}. Domain-specific pre-training can improve performance of downstream models on many tasks \citep{gururangan-etal-2020-dont}, including clinical \citep{alsentzer-etal-2019-publicly}, yet the impact of in-domain exposure is less studied when meta-evaluating faithfulness metrics. As such, we implement three versions of each metric with increasing levels of domain adaptation: \texttt{Off-The-Shelf} (fully out-of-domain), \texttt{Tuned In-Domain} (pre-trained out-of-domain, tuned-in-domain), and \texttt{Double In-Domain} (pre-trained and tuned in-domain). For in-domain pre-training, we rely on existing models pre-trained on clinical or biomedical corpora, specific to each dataset.  For in-domain metric tuning, we use the \texttt{Train - HIV} data from Table \ref{tab:data-stats}. Training details are provided as part of each metric description in \S \ref{sec:metrics}.

\paragraph{Output Lengths.} Given previous work \citep{adams-etal-2021-whats} detailing the lack of inter-sentence discourse markers in clinical narratives, we evaluate each sentence independently. Performing meta-evaluation of metrics at the sentence-level also increases the level of support ($29$ vs $245$) when computing instance-level correlations. This choice also enables us to explore the impact of sentence-level partitioning of summaries on metric performance.


\paragraph{Input Lengths.} Our inputs contain $\sim$ 30,000 tokens. Conditioning evaluation on the entire source is computationally expensive and often undesirable (e.g., entailment models are trained on short premises). Modern faithfulness metrics tend to struggle with long inputs \citep{honovich-etal-2022-true-evaluating}, likely due to the fact that only a handful of sentences from the source text are relevant to a given summary sentence \citep{lebanoff-etal-2019-analyzing}.

Yet, computing source-summary alignments \citep{ernst-etal-2021-summary} is particularly challenging for clinical text because 1) massive redundancy from copy-and-paste \citep{hirschtick2006copy}; 2) lexical variation in discussing semantically identical concepts (abbreviations, acronyms, etc.) \citep{adams2020zero}; 3) the need for complete context when assessing missing or misleading information. To explain 3), if a summary includes an outdated lab measurement, simply returning that single lab value as the alignment would provide a false sense of clinical correctness. The full chronology is needed.

Given this complexity, we separately evaluate the impact of alignment granularity (2-3 sentences to the whole input) on metric tuning and inference.

\begin{table}[h]
\centering
\small
\begin{tabular}{l|c}
\textbf{Alignment Method} & \textbf{Number of Source Sents} \\ \hline
\texttt{ROUGE-Gain} & 1.1 \\
\texttt{BS-Gain} & 1.8 \\
\texttt{ROUGE-TopK} & 5.0 \\
\texttt{BERT-TopK} & 5.0 \\
\texttt{Top Section} & 13.2 \\
\texttt{Entity Chain} & 15.3  \\
\texttt{Full} & 921.2* \\
\end{tabular}
\caption{The average number of source sentences aligned to each summary sentence for different alignment methods. $K$ is $5$. *\texttt{Full} differs for each metric based on token limits (pre-truncated lengths shown). } \label{tab:alignment-stats}
\end{table}

Each method aligns a summary sentence to a subset of sentences from the source. Duplicate source sentences are removed. Table \ref{tab:alignment-stats} shows the average number of aligned sentences by method.


\paragraph{Alignments - Granular.} \texttt{ROUGE-TopK} takes the $k=5$ highest ROUGE-aligned sentences (average of R1, R2, RL F-1), while \texttt{ROUGE-Gain} follows \citet{lebanoff-etal-2019-scoring} and maximizes the relative ROUGE gain of adding each additional sentence to the current set of aligned sentences. To account for lexical variation and noise, we also build alignments with BERTScore (BS) from in-domain weights (see description of BERTScore model used in \S \ref{sec:metrics}). \texttt{BS-TopK} selects the $k$ source sentences with the highest F-1 BS vis-a-vis the summary sentence. \texttt{BS-Gain} follows the approach in \citep{adams-etal-2022-learning} in which a coverage weight is assigned to each token in the summary sentence and updated based on the maximal alignment so far.

\paragraph{Alignments - Entity-Chain.} Given a summary sentence, we define an alignment method based on \texttt{Entity-Chains} \citep{barzilay1997using, narayan-etal-2021-planning} as the set of sentences in the source with at least one medical concept (a CUI from the Unified Medical Language System (UMLS) aligned to any of the CUIs in the summary sentence. Appendix \ref{sec:entity-extraction} details how entities are extracted, linked to the UMLS, and aligned. Alignment is based on manually annotating pairs of mentions and learning a light-weight classifier on features which include mention similarity (using contextualized from SapBERT \citep{liu-etal-2021-self}, TF-IDF overlap, and Levenshtein distance), CUI similarity (using a custom CUI2Vec model trained on MIMIC-III on our CUI vocabulary), and other UMLS-based features (TUI and semantic group).

\paragraph{Alignments - Section-Level.} To avoid fragmented alignments pulled from different notes, we also consider the Top-1 most aligned section as its own alignment. In particular, we select the section with the highest average ROUGE-\{1, 2, L\} overlap vis-a-vis each sentence in the summary.

\paragraph{Alignments - Full Input.} The conventional approach is to pass the whole source as input. Most of our inputs surpass both short and long transformer token limits. As needed for each metric, then, for \texttt{Full Input} alignments for each summary sentence, we select the source sentences with the highest ROUGE-\{1, 2\} overlap vis-a-vis summary sentence until a target token limit is reached.

\subsection{Metrics} \label{sec:metrics}

We describe each metric at a high-level and then detail domain adaptation. In Appendix \ref{app:other-metrics}, we introduce 2 additional metrics as part of the meta-evaluation: \texttt{ReDRESS} and \texttt{FactScore}, for which we only implement in-domain variants.


\paragraph{BERTScore.} \textbf{High-Level.} BERTScore \citep{zhang2019bertscore} computes a greedy soft-alignment, based on BERT hidden state similarities, between a reference and a hypothesis text. As in \citet{pagnoni-etal-2021-understanding}, we compute a \emph{reference-free} BERTScore: in our case, the hypothesis is a summary sentence and the reference its aligned source sentences. We only report BERTScore-Precision because it has the highest correlation with human judgment on our data. \textbf{Domain-Adaptation.} For \texttt{Off-The-Shelf}, we use \texttt{RoBERTA-Large}. There is no task-specific training for BERTScore so we report a single \texttt{In-Domain} variant. Specifically, we use a RoBERTA-Large model pre-trained from scratch with a custom BPE tokenizer on biomedical (PubMed and PubMed Central (PMC)), as well as clinical text (MIMIC-III) \citep{lewis-etal-2020-pretrained}\footnote{The model weights (RoBERTa-large-PM-M3-Voc-large) can be downloaded from \href{https://github.com/facebookresearch/bio-lm}{GitHub} and used with HuggingFace.} For all variants, given that alignments can exceed the RoBERTA context window of 512, we separately encode sentences from the same section and concatenate them (similarly to the paragraph chunking method from \citet{liu-lapata-2019-hierarchical}).

\paragraph{BARTScore.} \textbf{High-Level.} BARTScore \citep{yuan2021bartscore} computes the length-normalized log likelihood of a summary conditioned on the input. We measure BARTScore for each sentence based on its aligned source inputs. \textbf{Domain Adaptation.} For \texttt{Off-The-Shelf}, we use a BART-Large model fine-tuned on CNN/DailyMail news summaries\footnote{\texttt{facebook/bart-large-cnn} from HuggingFace.}. For \texttt{Tuned In-Domain} and \texttt{Double In-Domain}, we fine-tune BART-based models on \texttt{Train - HIV} corpus. The targets are single summary sentences and the inputs are their aligned source sentences. We fine-tune a separate model for each alignment method from \S \ref{sec:challenges}. For \texttt{Double In-Domain}, we initialize fine-tuning on \texttt{Train - HIV} with the BART-based ReDRESS model from \citet{adams-etal-2022-learning} \footnote{ReDRESS is pre-trained on a novel entity-based de-noising objective on unlabeled clinical text (MIMIC-III discharge summaries). The model weights are accessible on HuggingFace as ``griffin/redress-clinical-hallucination-generator''.}. For \texttt{Tuned In-Domain}, we initialize fine-tuning from \texttt{BART-Base} (to match ReDRESS). Using the Trainer from the Transformers library \citep{wolf-etal-2020-transformers}, we fine-tune each model in batches of $16$ for $10,000$ steps with a learning rate of $3e-5$ ($200$ warmup steps followed by linear decay). We use a label smoothing factor of $0.1$.

\paragraph{CTC.} \textbf{High-Level.} Compression, Transduction, Creation (CTC) \citep{ctc} defines a unified series of weakly supervised methods to evaluate system outputs on several NLG tasks. For summary faithfulness, the \texttt{CTC Score} represents the average number of tokens predicted as ``fake'' given the source. To train the CTC model, spans from reference summaries are masked-and-filled with a separate language model: the generator. \textbf{Domain Adaptation.} For \texttt{Off-The-Shelf}, we use \texttt{D-cnndm}, a RoBERTA-Large model fine-tuned for CTC consistency \textbf{d}iscrimination on the CNN/Dailymail dataset. For domain adapation, we corrupt summary sentences from \texttt{Train - HIV} and learn to discriminate based on source alignments. As in BARTScore, we fine-tune a separate discriminator for each alignment method from \S \ref{sec:metrics}. To generate fake tokens (the generator), we first train a mask-infiller (BART-base) on all discharge summaries in MIMIC-III. We use the same span mask procedure from CTC (based on a dependency parse) to align the training objective with its usage. We discuss generator training details and example outputs in Appendix \ref{app:ctc}. For \texttt{Double In-Domain}, we initialize the CTC Discriminator from the same biomedical RoBERTA model used for the \texttt{In-Domain} BERTScore \citep{lewis-etal-2020-pretrained}. For \texttt{Tuned In-Domain}, we initialize tuning from \texttt{RoBERTA-Large} (to match the initialization for \texttt{Off-The-Shelf}). We use the CTC codebase\footnote{\url{https://github.com/tanyuqian/ctc-gen-eval}} to train the discriminator with two modifications: we do not augment the data with paraphrasing\footnote{Existing paraphrase tools perform very poorly on clinical text and introduce many factual inconsistencies.}, and we train for 5 epochs (not 1).


\paragraph{Entailment.} \textbf{High-Level.} Faithful summaries should be entailed by the source text. \textbf{Domain Adaptation}. For \texttt{Off-The-Shelf}, we use a state-of-the-art entailment consistency model: SummaC \citep{laban-etal-2022-summac}. SummaC computes a faithfulness score for a summary sentence by computing separate entailment scores for each source-summary pair and then aggregating (either with a greedy argmax--as in BERTScore--in a zero-shot setting, or with a learned 1D convolution\footnote{\citet{falke-etal-2019-ranking} demonstrated that off the shelf NLI \citep{bowman-etal-2015-large} models, trained on sentence-to-sentence data, do not transfer well to summary faithfulness task (document-sentence(s))}. We use the latter: SummaC-Conv, which is tuned using news summary human annotations. For \texttt{In-Domain}, we do not have enough annotations on which to tune a SummaC-Conv model. Rather, we rely on the zero-shot setting, in which an off-the-shelf in-domain entailment model is used to score summary sentences. Specifically, we use the SciFIVE Model\footnote{The weights can be downloaded from the HuggingFace Transformers library via the following model card: \texttt{razent/SciFive-large-Pubmed\_PMC-MedNLI}.} with SOTA performance on the MedNLI dataset \citep{romanov2018lessons}--clinician-annotated entailment corpus whose premises come from MIMIC-III. SciFive is provided the summary sentence and its aligned source text as input, and generates a label: \texttt{\{contradiction, neutral, entailment\}}. For meta-evaluation, we convert each class label to an integer in the set $\{-1, 0, 1\}$. \\

\begin{figure*}[t]
\centering
\includegraphics[width=\linewidth]{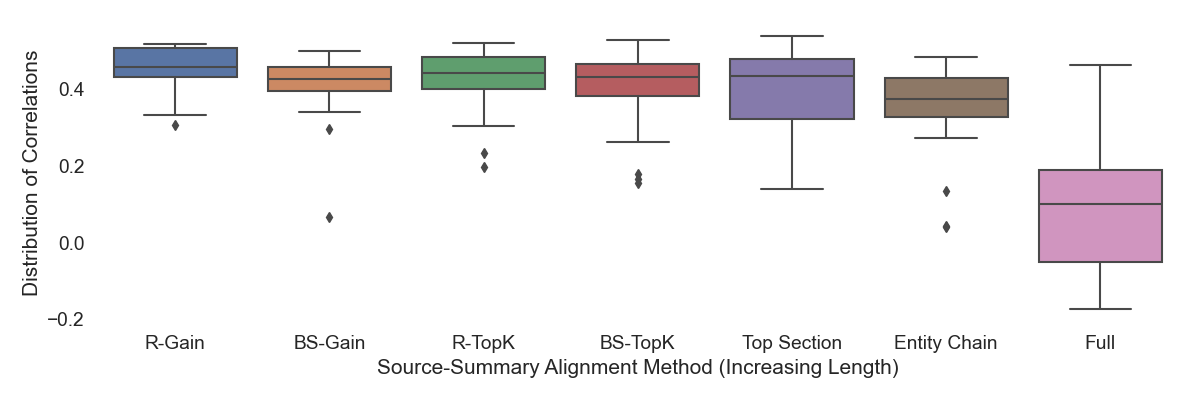}
\caption{The effect of alignment granularity on the distribution of instance-level Pearson correlations to human judgments across a wide range of metric variants (42). Correlations are more stable across metrics (higher average, higher minimum, and less overall variation) when the inputs (source-summary alignments) are shorter in length.} 
\label{fig:alignment-box}
\end{figure*}

\section{Meta-Evaluation of Metrics} \label{sec:meta}

Separately for each sentence of each summary in the human annotation set ($245$), we compute a human error rate \texttt{HErr}: defined as the fraction of summary elements (SE) in the sentence marked as either \texttt{Not In Notes}, \texttt{Incorrect}, or \texttt{Missing}. Unless explicitly stated, we do not distinguish between error type or severity (Minor, Critical) for the meta-evaluation. For the following analysis, we report the instance-level Pearson \citep{cohen2009pearson} correlation coefficient between \texttt{HErr} and metric scores (two $245$ length vectors).

\begin{table}[h]
\centering
\small
\begin{tabular}{l|c|cccc}
\multirow{2}{*}{\textbf{\texttt{Method}}} & \multirow{2}{*}{\texttt{\textbf{\makecell{\# \\ Sent}}}} & \multicolumn{4}{c}{\textbf{\texttt{Correlations}}} \\
& & \texttt{Avg} & \texttt{Max} & \texttt{Min} & \texttt{Std} \\ \hline
\texttt{ROUGE-Gain} & 1.1 & .46 & .52 & 0.31 & .06 \\
\texttt{BS-Gain} & 1.8 & .42 & .50 & .07 & .07 \\
\texttt{ROUGE-TopK} & 5.0 & .43 & .52 & .20 & .07 \\
\texttt{BERT-TopK} & 5.0 & .41 & .53 & .16 & .09 \\
\texttt{Top Section} & 13.2 & .40 & .54 & .14 & .10 \\
\texttt{Entity Chain} & 15.3 & .36 & .48 & .04 & .10  \\
\texttt{Full} & 921.2* & .09 & .46 & -.17 & .16 \\
\end{tabular}
\caption{ Average of instance-level correlation of metric scores to human correlations at the summary sentence-level. Each row represents an alignment method, which provides inputs of varying lengths to each metric, and corresponds to a column in the box plot in Figure \ref{fig:alignment-box}. } \label{tab:alignment-correlations}
\end{table}

\subsection{Finding the Optimal Source Granularity} \label{sec:sa}

\paragraph{Research Question.} How much of the source input (averaging $< 20k$ tokens across $> 40$ notes) is necessary to achieve high correlation with humans? 

\paragraph{Experimental Setup.} To answer this question, we vary the number of source sentences provided to \emph{every} metric and variant from \S \ref{sec:metrics} and analyze its impact on performance (instance-level Pearson correlation with the Human Error Rate, \texttt{HErr}).

\paragraph{Findings.} Figure \ref{fig:alignment-box} and Table \ref{tab:alignment-stats} reveal that, on average, metrics have higher correlations to human judgment when the inputs to the metric are shorter (with ROUGE-Gain being the shortest and having highest average Pearson Correlation of $.46$). The standard deviation of average instance-level correlations grows monotonically as alignments grow longer. Also, using the entire source is the most volatile (minimum of $-.17$) and the maximum correlation $.50$ is lower than the maximum correlation using a source-alignment (\texttt{Top Section}). These findings strongly suggest that scoring summaries based on the full source input is detrimental. 


\begin{table*}[ht]
\centering
\small
\begin{tabular}{cl|cccccc|c}
& &\multicolumn{6}{c}{\textbf{\texttt{Usage Alignment}}} & \\
& & \textbf{\texttt{R-Gain}} & \textbf{\texttt{BS-Gain}} & \textbf{\texttt{R-TopK}} & \textbf{\texttt{BS-TopK}} & \textbf{\texttt{\makecell{Top \\ Section}}} & \textbf{\texttt{\makecell{Entity \\ Chain}}} & \textbf{\texttt{\makecell{Tune \\ Avg}}} \\ \hline
\multirow{6}{*}{\textbf{\texttt{\makecell{Tune \\ Alignment}}}} & \textbf{\texttt{R-Gain}} & \cellcolor[gray]{0.9} \textbf{.467} & .449 & .458 & \underline{.449} & .397 & .344 & .427 \\
& \textbf{\texttt{BS-Gain}} & \textbf{.458} & \cellcolor[gray]{0.9} .387 & .427 & .382 & .396 & .351 & .400 \\
& \textbf{\texttt{R-TopK}} & \textbf{.449} & .440 & \cellcolor[gray]{0.9} .442 & .446 & .408 & \underline{.387} & .428 \\
& \textbf{\texttt{BS-TopK}} & \textbf{.460} & .411 & .435 & \cellcolor[gray]{0.9} .407 & .416 & \underline{.387} & .419 \\
& \textbf{\texttt{Top Section}} & \underline{\textbf{.469}} & .440 & .463 & .446 & \cellcolor[gray]{0.9} \underline{.427} & .379 & .437 \\
& \textbf{\texttt{Entity Chain}} & .452 & \underline{.450} & \underline{\textbf{.469}} & .438 & .407 & \cellcolor[gray]{0.9} .379 & .432 \\ \hline
& \textbf{\texttt{Usage Avg}} & .459 & .429 & .449 & .428 & .408 & .371 & \\
\end{tabular}
\caption{ Each row represents the Source-Summary alignments computed for metric \emph{tuning}, whereas the columns denote the alignment method for inference (\emph{usage}). Each cell represents the instance-level metric correlation to the Human Error Rate, averaged across four metric variants (BARTScore and CTC, \texttt{Tuned In-Domain} and \texttt{Double Domain}). The row-wise max is \textbf{bolded} and column-wise is \underline{underlined}. The diagonal is shaded in gray. } \label{tab:alignment-matrix}
\end{table*}

\subsection{Optimal Alignments for Metric Tuning} \label{sec:tuning-alignments}

\paragraph{Research Question.} \S \ref{sec:sa} reveals that shorter source alignments are preferable when \emph{using} metrics. Is the story the same when \emph{tuning} metrics? And should the alignment method used for metric tuning match the method used during inference?

\paragraph{Experimental Setup.} To answer this question, we breakdown metric performance (correlation to \texttt{HErr}) by the alignment method used for metric \emph{tuning} and, separately, for \emph{usage}. We consider 4 metrics (\texttt{Tuned In-Domain} and \texttt{Double In-Domain} variants for BARTScore and CTC). Each training instance is a summary sentence from \texttt{Train - HIV} and its aligned source context.

\paragraph{Findings.} Each cell in Table \ref{tab:alignment-matrix}\footnote{\texttt{Full} is not shown because it was not implemented for CTC due to token context restrictions for RoBERTA of 512.} represents an average of instance-level correlations to \texttt{HErr} across 4 metric variants (2 for BARTScore, 2 for CTC). Looking at the row-wise maximum values (\textbf{cells}), we notice that $5 / 6 $ involve using the shortest alignment (\texttt{R-Gain}) for metric \emph{usage}. This aligns with our analysis above in \S \ref{sec:sa}. Yet, the optimal alignment method for metric tuning is much less clear. If anything, we notice that $4 / 6$ of the column-wise maximum values (\underline{cells}) come from models tuned models from one of the two longest alignment methods (\texttt{Top Section} and \texttt{Entity Chain}). Additionally, on average, the diagonal values (shaded in gray) do not outperform the non-shaded regions. Taken together, at a high-level, this analysis suggests that additional context may be helpful when learning metrics (to make the task more difficult), yet, when using a metric, providing shorter, higher precision contexts are preferable.

\subsection{Effect of Summary Granularity} \label{sec:target-granularity}

\paragraph{Research Question.} For our meta-analysis, we measure faithfulness at the summary sentence level. As such, we have been scoring summaries sentence-by-sentence (\texttt{Sentence-Level}). Yet, for some metrics with localized predictions, alternatively, we can process the entire summary and then post-hoc extract sentence-level scores (\texttt{Summary-Level}). Which method leads to higher metric correlations?

We separately consider BARTScore and BERTScore to answer this research question. 

\begin{figure}[h]
\centering
\includegraphics[width=\linewidth]{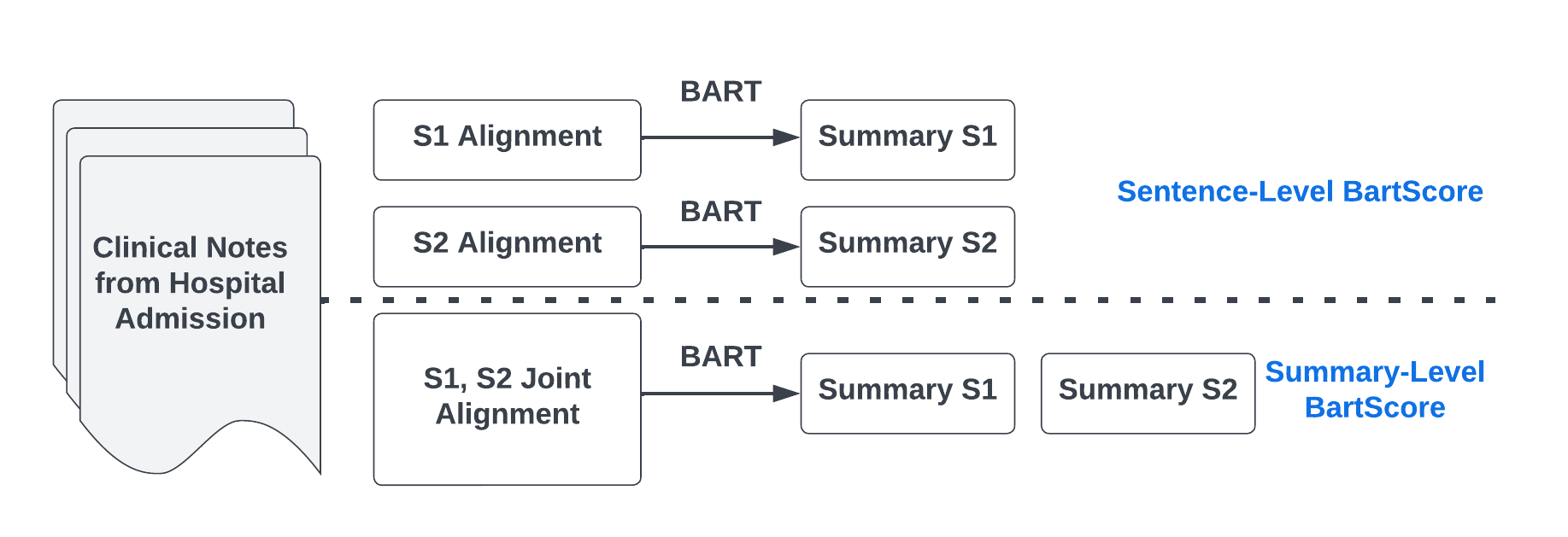}
\caption{ Sentence-Level BARTScore (BART-based) versus Summary-Level (LongFormer Encoder-Decoder (LED)). The LED scales BART to long inputs ($> 1024$ tokens). While Summary-Level generates a full summary, BARTScores are computed separately for each sentence by extracting logits from sentence boundaries.} 
\label{fig:bartscore-design}
\end{figure}

\paragraph{BARTScore Experimental Setup.} \texttt{Sentence-Level} is the default approach  for all metrics, as detailed in \S \ref{sec:metrics}. \texttt{Summary-Level} BARTScore involves processing the full summary conditioned on aligned source sentences. For this setting, we simply treat the summary as a ``single sentence'' and align it to the source sentences. Yet, these source alignments often exceed the BART context window ($1,024$ tokens). To handle longer inputs, we replace BART with an LED model (which scales up BART with sparse attention). We fine-tune for 10,000 steps on \texttt{HIV - Train}) (as in \texttt{Sentence-Level}) using the same LED hyper-parameters from Appendix \ref{app:led}. For both sets of experiments, we consider two alignment methods: \texttt{ROUGE-Gain} and \texttt{ROUGE-TopK}. $K = 5$ for \texttt{Sentence-Level} and for aligning to full summaries, $K = 300$. During inference, we pass the same alignment granularity on which the model was fine-tuned. \texttt{Summary-Level} and \texttt{Sentence-Level} are contrasted in Figure \ref{fig:bartscore-design}.

\begin{table}[h]
\centering
\small
\begin{tabular}{lc|c}
\multirow{2}{*}{\textbf{\texttt{\makecell{Summary \\ Granularity}}}} & \multirow{2}{*}{\textbf{\texttt{\makecell{Source \\ Alignment}}}} & \multirow{2}{*}{\textbf{\texttt{\makecell{Pearson \\ Correlation}}}} \\
& & \\ \hline
\multirow{2}{*}{\textbf{\texttt{\makecell{Summary \\ Level}}}} & ROUGE-Gain & .438 \\
 & ROUGE-TopK & .424 \\ \hline
\multirow{2}{*}{\textbf{\texttt{\makecell{Sentence \\ Level}}}} & ROUGE-Gain & .516 \\
& ROUGE-TopK & .481 \\
\end{tabular}
\caption{ BARTScore correlation to human faithfulness labels by summary granularity (processing the full summary at once as opposed to sentence-by-sentence). } \label{tab:target-granularity-bartscore}
\end{table}

\paragraph{BARTScore Findings.} Table \ref{tab:target-granularity-bartscore} reveals that \texttt{Sentence-Level} BARTScore (with separate alignments computed per sentence) is preferable to processing \texttt{Summary-Level} ($.516$ / $.481$ versus $.438 / .424$). This relates to the previous finding in \S \ref{sec:sa}. In both cases, tighter alignment between the inputs and outputs passed to a metric is preferable.

\begin{table}[h]
\centering
\small
\begin{tabular}{lc|c}
\multirow{2}{*}{\textbf{\texttt{\makecell{Summary \\ Granularity}}}} & \multirow{2}{*}{\textbf{\texttt{\makecell{Source \\ Alignment}}}} & \multirow{2}{*}{\textbf{\texttt{\makecell{Pearson \\ Correlation}}}} \\
& & \\ \hline
\textbf{\texttt{\makecell{Summary Level}}} & Full & .357 \\ \hline
\textbf{\texttt{\makecell{Sentence Level}}} & Full & .464 \\
\end{tabular}
\caption{ Correlation of BERTScore Precision to human labels by summary granularity (summary versus single-sentence). Both use the entire source (\texttt{Full} alignment). } \label{tab:target-granularity-bertscore}
\end{table}

\paragraph{BERTScore Experimental Setup.} We evaluate \texttt{In-Domain} BERTScore variants \citep{lewis-etal-2020-pretrained} which use the entire source (\texttt{Full} alignment method). Specifically, we compare our baseline BERTScore approach (\texttt{Sentence-Level}), which encodes each summary sentence independently, with a \texttt{Summary-Level} variant, which involves encoding the entire summary before computing a separate BERTScore for each sentence\footnote{Since we report BERTScore precision, we can compute the full similarity matrix before segmenting by sentence.}. The latter is typically how BERTScore is used.

\paragraph{BERTScore Findings.} Table \ref{tab:target-granularity-bertscore} shows that encoding sentences independently (\texttt{Sentence-Level}) leads to higher correlation with human assessments ($.46$ versus $.36$). Given how choppy clinical notes are, including neighboring sentences can add substantial noise to contextual embeddings of summary sentences.

\subsection{Curious Case of In-Domain Training} \label{sec:in-domain-results}

\paragraph{Research Question.} There is a wealth of evidence to demonstrate the beneficial impact of in-domain pre-training on clinical \citep{alsentzer2019publicly, lehman2023we} and biomedical \citep{gu2021domain} downstream tasks. Yet, to our knowledge, no previous work examines the benefits of in-domain pre-training on clinical evaluation metrics. Is domain adaptation: at the pre-training level, and at the task-specific fine-tuning level, necessary for developing clinical faithfulness metrics?

\paragraph{Experimental Setup.} We breakdown instance-level metric correlations by the level of domain adaptation: \texttt{Off-The-Shelf}, \texttt{Tuned In-Domain}, and \texttt{Double In-Domain}. We consider \texttt{BARTScore}, \texttt{CTC}, and \texttt{Entailment}\footnote{We report correlations for best performing variants with respect to the alignment method used for tuning and inference.}. Please see \ref{sec:metrics} for specific in-domain and out of domain weights used.

\begin{table}[h]
\centering
\small
\begin{tabular}{cl|c}
\texttt{\textbf{\makecell{Domain \\ Adaptation}}} & \texttt{\textbf{Metric}} & \textbf{\texttt{\makecell{Pearson \\ Correlation}}} \\ \hline
\multirow{4}{*}{\textbf{\texttt{\makecell{Off The \\ Shelf}}}} & BARTScore & .539 \\
 & CTC & .507  \\
 & Entailment & .453 \\
 & \texttt{\textbf{Average}} & \textbf{.501} \\ \hline
 \multirow{4}{*}{\textbf{\texttt{\makecell{Tuned \\ In-Domain}}}} & BARTScore & .522 \\
 & CTC & .462  \\
 & Entailment* & .450 \\
 & \texttt{\textbf{Average}} & \textbf{.478} \\ \hline
\multirow{3}{*}{\textbf{\texttt{\makecell{Double \\ In-Domain}}}} & BARTScore & .516 \\
 & CTC & .439  \\
  & Entailment* & .450 \\
& \texttt{\textbf{Average}} & \textbf{.468 } \\ \hline
\end{tabular}
\caption{ The impact of domain adaptation of metrics on correlation to human assessments. For in-domain ``Entailment*'', we use a model pretrained on biomedical text and fine-tuned on the MedNLI dataset. It is not tuned on our clinical text, so it does not neatly fit into either \texttt{Tuned In-Domain} or \texttt{Double In-Domain}.} \label{tab:in-domain}
\end{table}

\paragraph{Findings.} Table \ref{tab:in-domain} shows a curious trend: that increasing levels of metric domain adaptation is associated with lower correlation to faithfulness annotations at the metric-level and across systems (average declines $.501 \rightarrow .478 \rightarrow .468 $). Below, we link this outcome to summary extractiveness.

\begin{figure}[t]
\centering
\includegraphics[width=\linewidth]{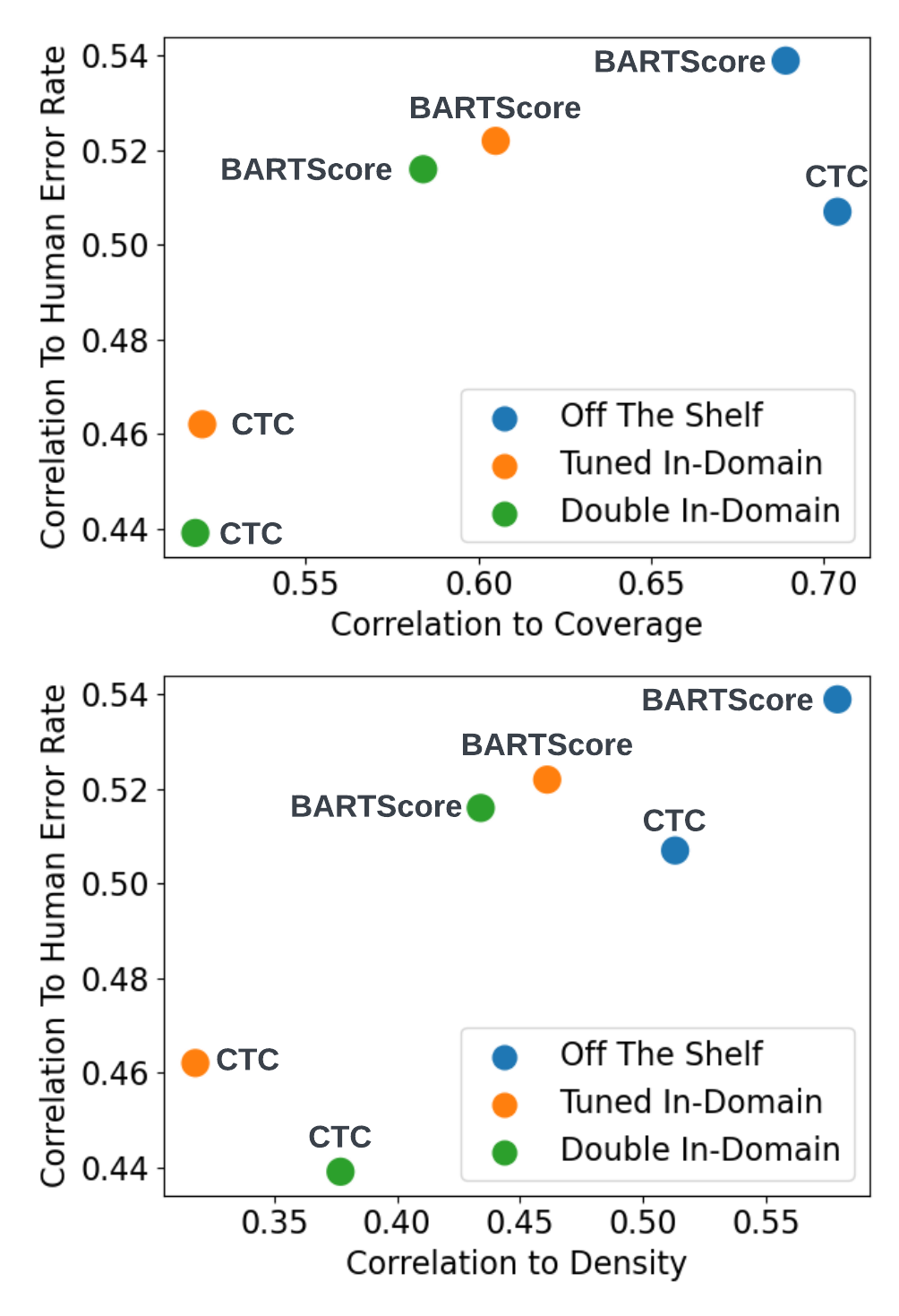}
\caption{Relationship between Correlation To Extractiveness and Correlation to Human Performance. Each dot represents the best performing (highest correlation) score across each source-summary alignment (see \S \ref{sec:sa}). }
\label{fig:spurious}
\end{figure}

\begin{figure*}[t]
\centering
\includegraphics[width=0.75 \linewidth]{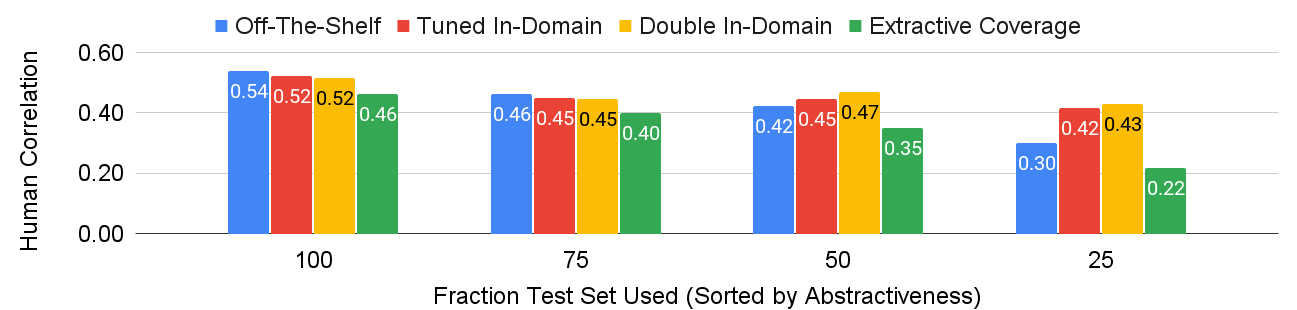}
\caption{Impact of summary extractiveness on metric correlation to human labels. BARTScore variants with different levels of in-domain training are shown, along with Extractiveness (Coverage). Coverage shows the steepest decline in correlation to human labels as average coverage declines, followed by the BARTScore variant most correlated to it (\texttt{Off-The-Shelf}). Metrics with in-domain training perform best on the more abstractive subsets.}
\label{fig:density}
\end{figure*}

\begin{figure*}[t]
\centering
\includegraphics[width=0.75 \linewidth]{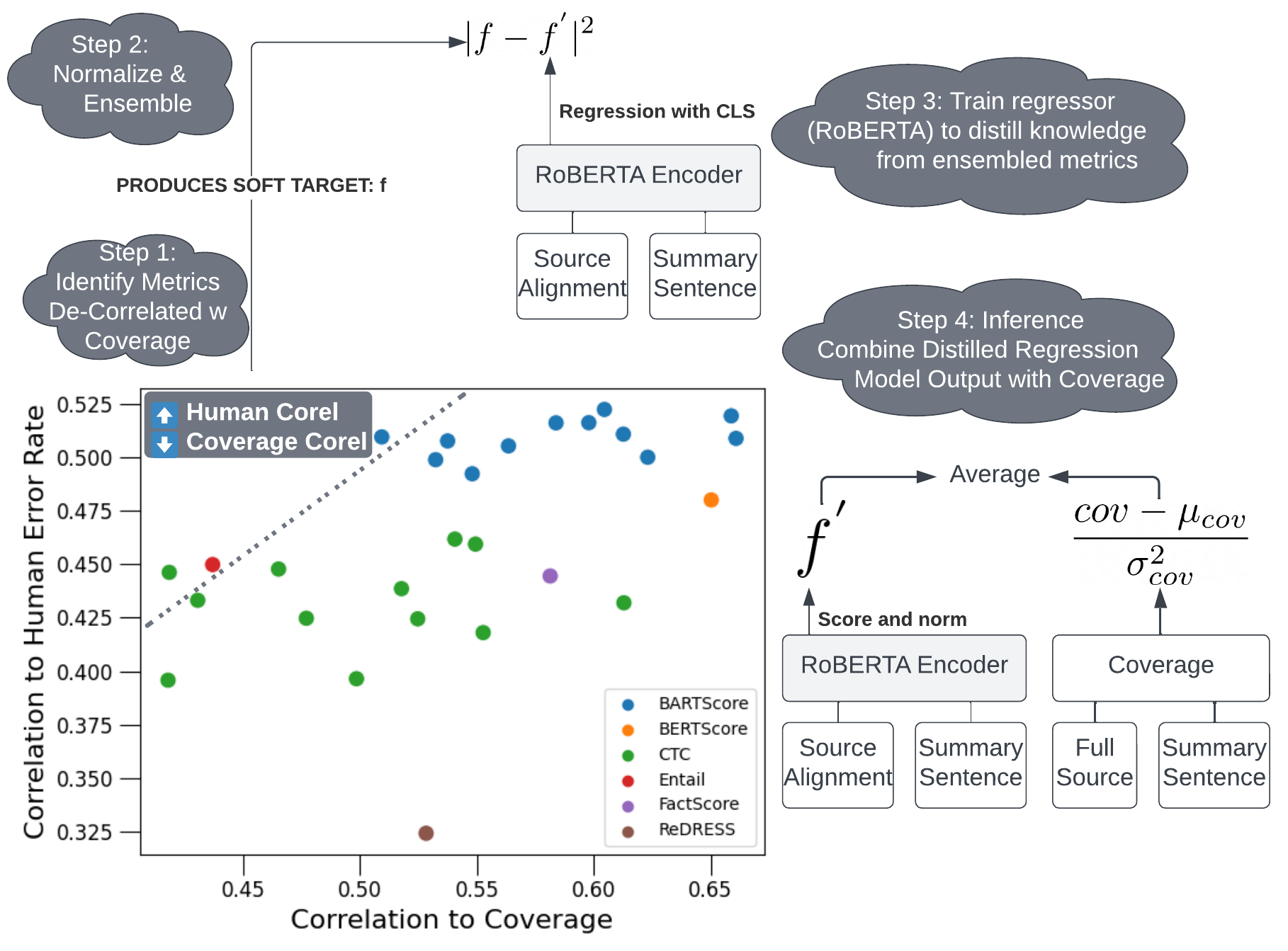}
\caption{\textbf{Step 1}: Identify Optimal Metrics for Knowledge Distillation: High Correlation to Human Labels and Low Correlation to Extractive Coverage. \textbf{Step 2}: Normalize and ensemble (average) to produce produce soft targets $f$ on the \texttt{Train - HIV} cohort. \textbf{Step 3}: Train a model (RoBERTA) as a regressor ($f^{'}$) against the ensembled soft targets $f$. \textbf{Step 4}: Create a combined metric: \texttt{\textbf{Distilled + Coverage}}, which combines the score from the RoBERTA model--distilled from metrics relatively less correlated with coverage--with a normalized coverage score. } 
\label{fig:distillation-metric}
\end{figure*}

\paragraph{Spurious Correlates Hypothesis.} \citet{durmus-etal-2022-spurious} find that reference-free metrics over rely on spurious correlates: variables which are highly correlated to human annotations on a biased test set, yet less correlated on a more realistic, diverse data distribution. Identifying such correlates is important because it suggests a metric brittleness which may not be captured by simple correlation analysis. As in their work, we focus on summary extractiveness \citep{grusky-etal-2018-newsroom} as the potentially spurious correlate. In Figure \ref{fig:spurious}, we reveal a clear pattern between metric correlation to extractiveness and correlation to the human error rate. In particular, across Coverage (top) and Density (bottom), high correlations to extractiveness are positively related to the correlation with the human error rate. Additionally, we see that in-domain training de-correlates metrics to extractiveness (\texttt{Tuned-In-Domain} and \texttt{Double In-Domain}. To examine why this might be the case, we examine the extractiveness of reference versus system summaries and a clear bias emerges.

\begin{table}[h]
\centering
\small
\begin{tabular}{l|cc}
\textbf{\texttt{Summary}} & \textbf{\texttt{Coverage}} & \textbf{\texttt{Density}} \\ \hline
\textbf{Reference} & 0.88 & 12.04 \\
\textbf{Model-Generated} & 0.95 & 39.12 \\
\end{tabular}
\caption{ Model-Generated summaries are \emph{substantially} more extractive (Coverage, Density) than the references on which they are trained. This creates a train-test mismatch for metrics, which are fine-tuned on abstractive summaries and meta-evaluated on extractive ones. } \label{tab:extractiveness-mismatch}
\end{table}

Table \ref{tab:extractiveness-mismatch} shows that references are substantially more extractive in terms of both coverage (percentage of unigrams copied from the source) and density (average squared length of copied fragments) \citep{grusky-etal-2018-newsroom}. In other words, clinicians write more abstractive summaries than the Longformer. To more closely approximate more abstractive, clinician-authored summaries, we examine changes in correlations to human judgments as we filter for more abstractive subsets of the test set.  We sort system summary sentences in the test set by coverage and filter for smaller and smaller subsets (making the average coverage lower). Figure \ref{fig:density} reveals that in-domain BARTScore metrics start to outperform when summaries are more abstractive ($.30 \rightarrow .42 \rightarrow .43$ for the smallest bucket, i.e., the top 25\% most abstractive sentences in the eval set).

\paragraph{Domain-Adapted Metrics are Complementary to Coverage.} Recent work demonstrates the efficacy of ensembling de-correlated metrics \citep{kasai-etal-2022-bidimensional, colombo2022glass}. In light of our previous analysis, we can normalize each metric variant from Figure \ref{fig:spurious} and ensemble it with a normalized score for extractiveness (e.g., coverage). To make this explicit, given raw metric score $f$ and raw coverage $cov$, we create a combined metric $g$

$$
g = \frac{1}{2} * \left( \frac{f - \mu_{f}}{\sigma_{f}^{2}} + \frac{cov - \mu_{cov}}{\sigma_{cov}^{2}} \right)
$$

\noindent where $\mu$ and $\sigma$ represent mean and standard deviations for $f$ and $cov$ across all summary sentences. We can then insert each metric in Figure \ref{fig:spurious} as $f$ into this equation and compare correlations to \texttt{HErr}.

\begin{table}[h]
\centering
\small
\begin{tabular}{cl|c}
\texttt{\textbf{\makecell{Domain \\ Adaptation}}} & \texttt{\textbf{Metric}} & \textbf{\texttt{\makecell{Pearson \\ Correlation}}} \\ \hline
 & Coverage (Cov) & .457 \\ \hline
\multirow{4}{*}{\textbf{\texttt{\makecell{Off The \\ Shelf}}}} & BARTScore + Cov & .542 \\
 & CTC + Cov & .522  \\
 & Entailment + Cov & .524 \\
 & \texttt{\textbf{Average}} & \textbf{.529} \\ \hline
 \multirow{4}{*}{\textbf{\texttt{\makecell{Tuned \\ In-Domain}}}} & BARTScore + Cov & .547 \\
 & CTC + Cov & .523  \\
 & Entailment + Cov & .535 \\
 & \texttt{\textbf{Average}} & \textbf{.535} \\ \hline
\multirow{4}{*}{\textbf{\texttt{\makecell{Double \\ In-Domain}}}} & BARTScore & .547 \\
 & CTC + Cov & .514 \\
  & Entailment + Cov & .535 \\
& \texttt{\textbf{Average}} & \textbf{ .532 } \\ \hline
\end{tabular}
\caption{ The impact of domain adaptation on metric correlation to human assessments when combining with an easy-to-compute extractiveness statistic (coverage).  } \label{tab:plus-coverage}
\end{table}

Table \ref{tab:plus-coverage} reveals that when combining metrics with coverage, In-Domain adaptation slightly helps. \texttt{Off-The-Shelf} averages across three metrics (+ Cov) are $.529$ versus $.535$ and $.532$ for \texttt{Tuned In-Domain} and \texttt{Double In-Domain}, respectively. Yet, the differences are still relatively minor. 

\paragraph{Adapting to System Outputs with Knowledge Distillation.} Despite modest gains, domain adaptation does not help much, which may be due in part to differences in reference summaries versus system outputs. The above metrics are all trained solely on gold-standard references yet meta-evaluated on system outputs. To bridge this gap, we can learn a metric from system outputs. Yet, our annotation set is too small to use for this task. 

Instead, we leverage the fact that metrics, when ensembled, achieve relatively high correlation with human judgments, to create soft pseudo-targets on a larger set of system outputs (from \texttt{Train - HIV}). The goal, then, is to distill a single metric from the combined ``knowledge'' of multiple metrics\footnote{This kind of distillation is distinct yet related to conventional knowledge distillation \citep{hinton2015distilling}, which typically involves using a large teacher to train a smaller student.}. To do this, we first generate summaries with our LED model on the \texttt{Train - HIV} subset and segment into sentences. To produce pseudo targets, as shown in Figure \ref{fig:distillation-metric}, we identify a subset of In-Domain metrics with desired attributes: high-correlation to human labels and relatively low correlation to coverage. We then score each summary sentence with each metric in the ensemble, normalize the scores on a per-metric basis, and then average them to produce 
 pseudo-target $f$ for each training instance. We then train a student model, which receives as input a concatenation of a model-generated summary sentence and its aligned source context, and outputs a scalar: $f'$ using the \texttt{[CLS]} hidden state. The student is trained with a standard MSE loss: $|f' - f|^{2}$ and is initialized from clinical/biomedical RoBERTA \citep{lewis-etal-2020-pretrained}. We train in batches of $8$ for $10,000$ steps with a learning rate of $1e-5$ ($200$ warmup steps, followed by linear decay). For usage, we can \emph{optionally} combine the distilled score with the coverage score.

Via distillation of metrics which are relatively de-correlated with coverage, the goal is two-fold: to learn a single model that achieves a higher correlation on its own to other single-metric variants, and is complementary to coverage when combined.

\begin{table}[h]
\centering
\small
\begin{tabular}{l|c}
\textbf{\texttt{Metric}} & \textbf{\texttt{\makecell{Pearson \\ Correlation}}} \\ \hline
\textbf{Best Single Metric} & .539 \\
\textbf{Best Single Metric + Cov} & .547 \\
\textbf{Distilled Metric} & .564 \\ 
\textbf{Distilled + Cov} & \textbf{.573} \\
\end{tabular}
\caption{ Distilling a metric from the subset of metrics which are relatively less correlated to extractiveness (coverage) yields higher correlation with human labels than any other single metric. Additionally, combining the distilled metric with ($+$ Cov) obtains yields superior correlations to all single metric + coverage variants. } \label{tab:distillation}
\end{table}

Table \ref{tab:distillation} reveals that the Distilled metric outperforms the best baseline metric variant ($.564$ vs $.539$) and, because it is distilled from metrics which are relatively de-correlated with coverage, can be combined at inference with coverage to achieve an even higher correlation ($.573$). We ran a one-sided Williams Test \citep{graham-baldwin-2014-testing} to estimate the significance of increase in correlation to human labels from \texttt{Best Single Metric + Cov} to \texttt{Distilled + Cov}. The p-value was $.081$. As such, we cannot state that the impact of distillation is statistically significant at $p < 0.05$. But, we note that the sample size is small ($245$).

\begin{table}[h]
\centering
\small
\begin{tabular}{l|cc} 
\multirow{2}{*}{\textbf{\texttt{Metric}}} & \multicolumn{2}{c}{\textbf{\texttt{Pearson Correlation}}} \\ & \textbf{\texttt{\makecell{Single}}} & \textbf{\texttt{\makecell{Avg In Ensemble}}} \\ \hline
\textbf{Coverage (Cov)} & .457 & .544 \\ \hline
\textbf{BARTScore} & .539 & .550 \\
\textbf{CTC} & .507 & .546 \\
\textbf{Entailment} & .453 & .539 \\
\textbf{BERTScore} & .482 & .535 \\
\textbf{Reviser} & .324 & .528 \\
\textbf{FactScore} & .444 & .536 \\ \hline
\textbf{Distilled} &  \bf .564 & .556 \\ \hline
\textbf{Best Ensemble} & N/A & \bf .583 \\
\end{tabular}
\caption{ Comparing the correlation to human annotations of single metrics, as well as the average correlation of ensembles of metrics that include a given metric. Lastly, we include the correlation of the best performing ensemble of metrics (Coverage, BARTScore, Distilled). } \label{tab:ensemble}
\end{table}

\paragraph{Multi-Metric Ensembles.} Previously, we reported promising performance of our proposed Distilled metric--both on its own and combined with an extractiveness statistic. Yet, ideally, we would also want a metric that improves correlation when ensembled with other metrics. To this end, we enumerate all possible ensembles from a set which includes the coverage statistic and 7 metrics: our distilled model and our 6 implemented metrics (BARTScore, BERTScore, CTC, Entailment, FactScore, ReDRESS)\footnote{We report the best performing variant across in-domain pre-training / tuning and source-summary alignment methods.}. This provides us with $\sum_{n=1}^{N=8}{N \choose n} = 255$ unique ensembles, of which each metric takes part in $128$. Table \ref{tab:ensemble} shows correlation of metrics to \texttt{HErr} for metrics on their own (\texttt{Single}), as well as the average correlation to \texttt{HeRR} for metric ensembles which include a given metric (\texttt{In Ensemble}). Firstly, the metric rankings induced by \texttt{Single} and \texttt{In Ensemble} are mostly in agreement. Distilled outperforms all baselines on its own ($.564$) as well as its average correlation when used in an ensemble ($.556$). The last row of Table \ref{tab:ensemble} shows the correlation of the ensemble with the highest correlation to \texttt{HErr}: Coverage, BARTScore, and Distilled. To test significance of the \texttt{In Ensemble} results, we bootstrap $95\%$ confidence intervals (CI) for each metric's average \texttt{In Ensemble} correlation ($1000$ samples with replacement from vectors of size $128$) and find that the average correlation when \texttt{Distilled} is a part of an ensemble is significantly higher ($p < 0.05$) than the average correlation of any of the other 6 metrics (when part of an ensemble).

These results demonstrate that \texttt{Distilled} is useful on its own and is complementary to other metrics. More broadly speaking, the relative out-performance of ensembling (\texttt{In Ensemble} over \texttt{Single}) supports the notion that, when developing a metric, it is more useful to focus on its complementarity to existing metrics, rather than its performance in isolation~\citep{colombo2022glass}.

\subsection{Correlation by Metric Type} \label{sec:meta-by-type}

Previously, we meta-evaluated metrics against the percentage of summary elements (SE) with \emph{any} error. In this section, we breakdown metric correlations separately by error category: \texttt{Incorrect}, \texttt{Missing}, and \texttt{Not in Notes}.  We analyze metrics at the sentence-level against the percentage of Summary Elements in the sentence marked with a certain error. To provide more granular insights, we breakdown error type correlations by Domain Adaptation, Source-Summary Alignment methods, and metric classes (BARTScore vs CTC, etc).

\begin{figure}[h]
\centering
\includegraphics[width=\linewidth]{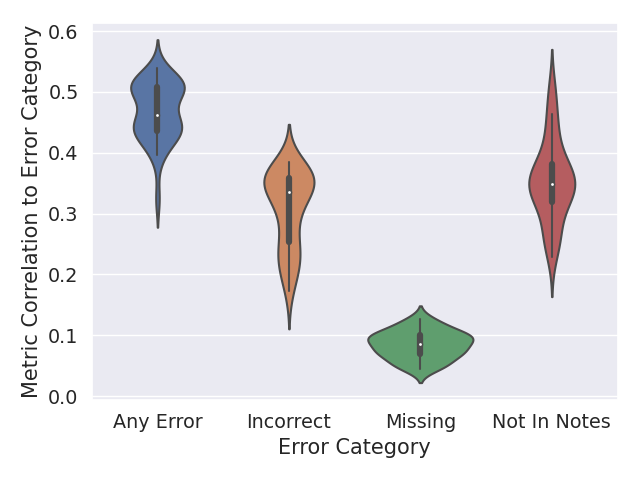}
\caption{Distribution of Metric Correlations to Human annotations by Category (includes Minor and Critical). } 
\label{fig:by-category}
\end{figure}

\begin{figure}[h]
\centering
\includegraphics[width=\linewidth]{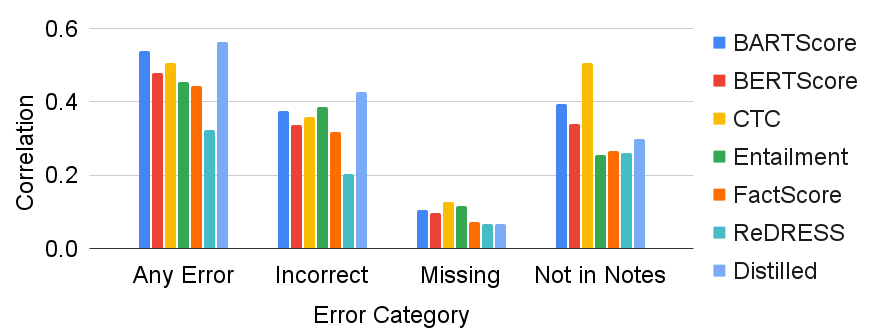}
\caption{Metric Correlations to Human Judgments by Error Category for each class of metrics from \S \ref{sec:metrics}. } 
\label{fig:by-metric}
\end{figure}

Figure \ref{fig:by-category} shows that \texttt{Missing} is the hardest for metrics (the instance-level correlations of metrics to fraction Missing across metric variants), which makes sense given its negligible correlation with Coverage ($.021$). Not in Notes are the simplest as they tend to be most associated with lexical overlap: $.391$ Pearson correlation between coverage and fraction of SE's in a sentence identified as \texttt{Not in Notes}. \texttt{Incorrect} errors can be subtle and are less correlated to coverage than \texttt{Missing}: $.249$. More generally, the over-reliance of these metrics on the level of copy-and-paste obfuscates their actual ability to reason over clinical narratives.

\paragraph{Metric-Wise.} Figure \ref{fig:by-metric} breaks down correlations to human judgments by metric and errory category. The primary take-away is that metric performance (here, correlation) does not exhibit monotonicity across error categories. Excluding Distilled, BARTScore is best at identifying \texttt{Any Error}, while Entailment outperforms on \texttt{Incorrect Errors}, and CTC performs best on \texttt{Not in Notes}. As discussed before, all metrics perform poorly on identifying missing content. CTC learns to identify extrinsic hallucinations so its strong performance on \texttt{Not in Notes} makes sense. Entailment metrics are trained on NLI datasets, which target the kinds of logic and inconsistency errors found in \texttt{Incorrect}. All metrics struggle with \texttt{Missing}. Taken together, these findings reveal that there is no one-size fits all solution to evaluation and we believe that metrics should be designed to fit the particular needs of a system and dataset \citep{pagnoni-etal-2021-understanding}. Reporting a single score for meta-evaluation obscures important differences across categories, as well as ignores the potential complementarity of different metrics. Given the potential of ensembling, targeted metrics--which out-perform on one category--may be more valuable to real-world use cases than ``jack of all trades, master of none''-type metrics.

\section{Conclusion} \label{sec:conclusion}

We collect fine-grained faithfulness annotations of Hospital Course summaries from clinicians and benchmark metrics against them. For each metric, we consider dimensions relevant to long-form clinical summarization: domain adaptation, input lengths, and output lengths. We find that metrics over-rely on the level of copy-and-paste in summaries. We can exploit this by computing a score which combines normalized extractiveness (coverage) with a new metric, which is distilled from a subset of the metrics most de-correlated with coverage. Moreover, metrics struggle with errors which require deep clinical knowledge (such as missingness, identification of mistakes from the source notes, etc.). While semi-supervised learning from synthetic datasets could help,  learning from explicit human feedback will likely be necessary for deployment in real-world, human-in-the-loop clinical settings in order to more tightly align metric behavior with clinical reasoning \citep{wei2022chain}.

\section*{Acknowledgements}

This research was supported by the National Library of Medicine (NLM) and National Institute of Allergy and Infectious Diseases (NIAID) of the National Institutes of Health (NIH) under Award Number T15LM007079. The content is solely the responsibility of the authors and does not represent the official views of the NIH.

\bibliography{anthology,custom}
\bibliographystyle{acl_natbib}

\appendix

\section{LED Training Details} \label{app:led}

\paragraph{Coarse Filtering.} The average length of the inputs ($\sim$ 30,000 tokens) exceeds the maximum sequence length even for transformer models with sparse attention mechanisms designed for long input sequences \citep{dai2019transformer, zaheer2020big, guo2021longt5}. Similarly to \citet{liu-lapata-2019-hierarchical}, we learn a simple bi-LSTM model which learns the relevance of each section, to predict the average ROUGE-1 and ROUGE-2 recall of each section vis-a-vis the reference.  In particular, we pass a bi-LSTM over the tokens in each section and compute a soft cross-entropy loss between the gold-standard ROUGE-2 recall and the predicted logit (sigmoid(score)). Then, we score each section and filter for the top-K sections. The top 100 sections are provided by an oracle during training and by the model for evaluation.

\paragraph{Fine-Tuning.} We fine-tune the Longformer Encoder-Decoder (LED) for 10 epochs with a batch size of 1 and gradient accumulation steps of 16.  We set the maximum learning rate to $3e-5$ (tuned in range the range of $1e-6$ to $1e-3$) with a warmup of $200$ steps with linear decay. The maximum input size was set to 16,384 and outputs were produced with minimum length of 64, maximum length of 1,024, trigam-blocking, and a beam size of 4 with length penalty 4.0. Training took 8 days on 1 NVIDIA RTX 3090 GPU (24GB).

\section{Entity Extraction} \label{sec:entity-extraction}

We extract and link entities to the Unified Medical Language System (UMLS \citep{bodenreider2004unified}) with CLAMP \citep{soysal2018clamp} and embed each entity mention with SapBERT \citep{liu-etal-2021-self} and first merge all entity mentions which share the same CUI from the UMLS. Exact match of two entities by CUI is far too strict given the size of the UMLS vocabulary as well as extraction noise from abbreviations, acronyms, etc. \citep{adams2020zero}. Then, we treat two distinct CUIs as synonyms based on a random forest classifier. The authors of this paper manually labeled 1,000 pairs of entities sampled from 10 different admissions, from a held-out set. The labels were \texttt{Unrelated}, \texttt{Related}, \texttt{Synonyms}. \textit{Ceftriaxone} is \texttt{Related} to \textit{antibiotics} since it is in the class of antibiotic, while it is a synonym of \textit{Rocephin}, its brand name. We split the 1,000 manually labeled examples into an 80-20 train-test split and compute features for all pairs of unique CUIs. They include similarity scores (cosine similarity) between CUIs, where CUI embeddings are provided by a pre-trained section-level CUI2Vec model on our corpus, as well as maximum pairwise alignments between mentions from different CUI sets: cosine similarity between SapBERT mention embeddings and lexical similarity (IDF overlap and string levenshtein distance), and finally, binary indicators for TUI and semantic group status from the UMLS.


\begin{figure*}[t]
\centering
\includegraphics[width=\linewidth]{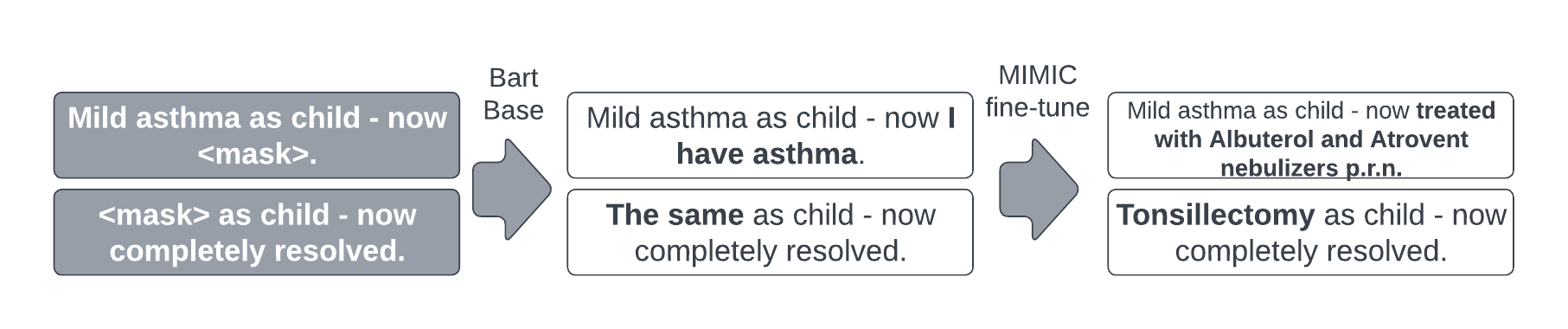}
\caption{The improvement in Mask-And-Fill completions after fine-tuning in-domain (MIMIC-III Discharge summaries) for just 500,000 steps. Syntactic spans are masked according to the procedure in \citet{ctc}. } 
\label{fig:clinbart-example}
\end{figure*}

\section{CTC Generator Details} \label{app:ctc}

 We use the same masking procedure used to train the CTC model to align the pre-training with the use case and use a BART-Base model to train for 500,000 steps with a batch size of $50$ and maximum learning rate of $2.2e-4$, linearly decaying after $200$ warmup steps. We show an example of the improvement in Mask-Infilling in Figure \ref{fig:clinbart-example}. 

 \section{Other In-Domain Metrics} \label{app:other-metrics}

 \paragraph{ReDRESS.} ReDRESS \citep{adams-etal-2022-learning} uses a novel hybrid approach that incorporates entity-swapping into a de-noising framework to generate synthetic corruptions on clinical text. Contrastive learning is used to teach another model to reverse the synthetic hallucinations. We adapt it as a faithfulness metric by revising model outputs conditioned on aligned source context and then measuring the revision intensity, e.g., how much was each summary edited to become faithful. We return the BERTScore F-1 between revised and un-revised summaries as the \texttt{ReDRESS-Score}: a higher score suggests fewer edits are necessary to re-write the summaries such that they are faithful.

\paragraph{FactScore.} As in \citet{adamsdesired}, FactScore is based on the state of the art model (MultiVERS \citep{wadden-etal-2022-multivers}) trained on the SciFact dataset \citep{wadden-etal-2020-fact}. SciFact is an expert-annotated dataset of 1,409 sentence-level scientific claims. Each summary sentence is scored conditioned on its aligned source sentences (which are varied according to the methods described in \S \ref{sec:challenges}). The \texttt{FactScore} is the probability that the MultiVERS assigns to the \texttt{SUPPORTED} label.

 \begin{figure*}[t]
\centering
\includegraphics[width=\linewidth]{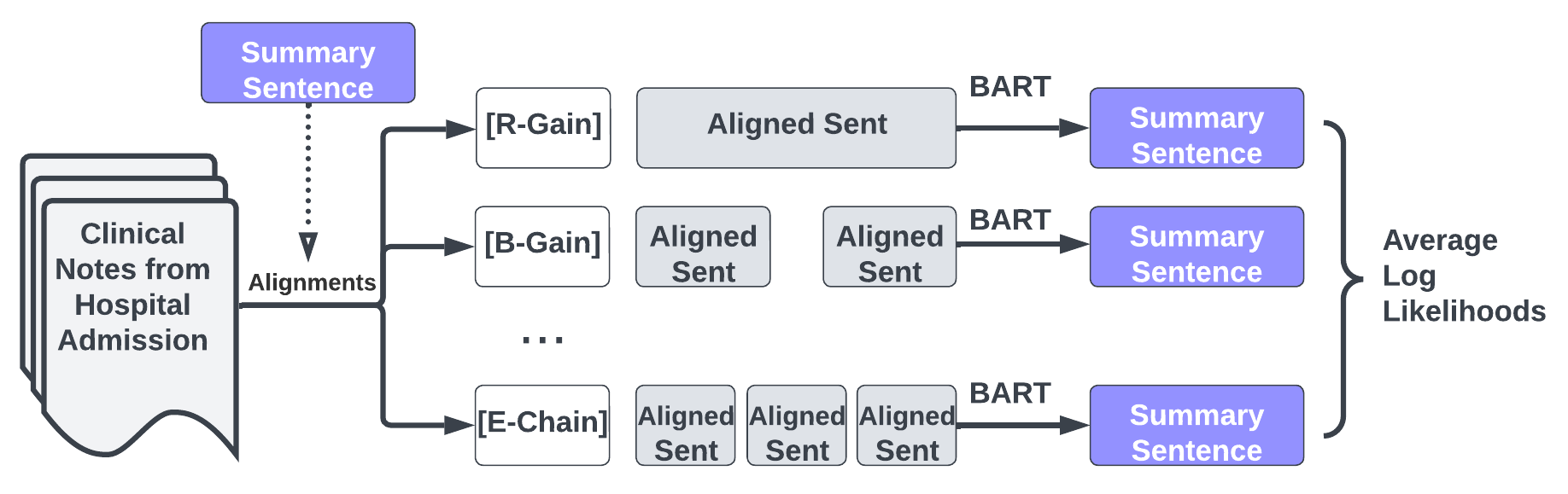}
\caption{MACS: \emph{M}ixture of \emph{A}lignment \emph{C}ontrolled BART\emph{S}cores.  We fine-tune a BARTScore model on augmented dataset, which creates a training instance for each source-summary sentence alignment (across 6 different alignment methods). Then, during inference, the model receives each alignment separately, along with an alignment-specific prefix embedding. The alignment-specific BARTScores are then averaged to produce a single ensembled score.  } 
\label{fig:macs}
\end{figure*}

\begin{table}[h]
\centering
\small
\begin{tabular}{l|c}
\textbf{\texttt{Source-Alignment}} & \textbf{\texttt{Pearson Correlation}}  \\ \hline
ROUGE-Gain & .516 \\
BERT-Gain & .427 \\
ROUGE-TopK & .481 \\
Top Section & .499 \\
Entity Chain & .380 \\ \hline
Average - Single Method & .461 \\ \hline
Alignment-Controlled Mixture & .496 \\
\end{tabular}
\caption{ Post-Hoc combination of BARTScores, where a separate encoder-decoder pass is made with inputs of variable length (based on different methods to compute source-summary alignments). } \label{tab:macs}
\end{table}

\section{\emph{M}ixture of \emph{A}lignment \emph{C}ontrolled BART\emph{S}cores} \label{app:macs}

\paragraph{Complementarity of Alignments.} In a sense, each source-summary alignment can be viewed as its own \emph{hard} attention head.  Inspired by the success of Multi-Head Attention \citep{vaswani2017attention}, as well as the success of ensembling de-correlated metrics \citep{kasai-etal-2022-bidimensional}, we propose a new BARTScore variant, MACS: ``\emph{M}ixture-of-\emph{A}lignment-\emph{C}ontrolled-BART\emph{S}cores''. At a high-level, we fine-tune a single BART model on our \texttt{Train - HIV} examples, in which the dataset consists of source alignment-summary sentence pairs (using each algorithm in \S \ref{sec:challenges}). In other words, for each summary sentence, we create 6 training instances: the input is a special code indicating the alignment, the aligned source text, and the target output is the summary sentence. Then, during inference, we separately compute a BARTScore for each alignment and average the BARTScores (mean token-level log-likelihood of generating a predicted sentence). We show the procedure for MACS in Figure \ref{fig:macs}. Table \ref{tab:macs} demonstrates BARTScore variants trained \emph{and} evaluated on different alignments. The results show that while our proposed extension--Alignment-Controlled Mixture, outperforms the average single alignment correlation to human annotations ($.496$ versus $.461$), it does not outperform the top performing single alignment method: \texttt{ROUGE - Gain} ($.496$ versus $.516$). We leave more sophisticated mixing for future work, which could involve learning a dynamic mixture during training \citep{lewis2020retrieval} or enforcing self-consistency \citep{wang2022self} on the decoder across each alignment.

\end{document}